\pdfoutput=1

\documentclass[11pt]{article}

\usepackage{acl}

\usepackage{times}
\usepackage{latexsym}
\usepackage{amsmath} 
\usepackage{placeins}

\usepackage[T1]{fontenc}
\usepackage[utf8]{inputenc}

\usepackage{microtype}

\usepackage{inconsolata}

\usepackage{graphicx}
\usepackage{booktabs}
\usepackage{lipsum}
\usepackage{multicol}
\usepackage{multirow}
\usepackage{color,xcolor,colortbl}
\usepackage{paralist}
\usepackage{enumitem}
\usepackage{subcaption}

\usepackage[ruled,vlined,linesnumbered]{algorithm2e}
\usepackage{hyperref}
\usepackage{cleveref}
\usepackage{todonotes}
\usepackage{float}

\newcommand{\ctrue}{\mathrm{T}}
\newcommand{\cfalse}{\mathrm{F}}

\newcommand{\xv}{\mathbf{x}}
\newcommand{\yv}{\mathbf{y}}

\newcommand{\multirowcell}[1]{\begin{tabular}[c]{@{}c@{}}#1\end{tabular}}

\newcommand{\ecoparagraph}[1]{\vspace{0.1cm}\noindent\textbf{#1}}
\newcommand{\conf}{C}

\title{Unconditional Truthfulness: 
Learning Unconditional Uncertainty  \\
of Large Language Models
}

\author{
\bf Artem Vazhentsev\textsuperscript{2,3}\quad
Ekaterina Fadeeva\textsuperscript{4}\quad
Rui Xing\textsuperscript{1,5}\\
\bf Gleb Kuzmin\textsuperscript{3,6}\quad
Ivan Lazichny\textsuperscript{3}\quad
Alexander Panchenko\textsuperscript{2,3}\quad
Preslav Nakov\textsuperscript{1}\\
\bf Timothy Baldwin\textsuperscript{1,5}\quad
Maxim Panov\textsuperscript{1}\quad
Artem Shelmanov\textsuperscript{1}\\
\textsuperscript{1}MBZUAI\quad
\textsuperscript{2}Skoltech \quad
\textsuperscript{3}AIRI \quad
\textsuperscript{4}ETH Zürich\\
\textsuperscript{5}The University of Melbourne\quad 
\textsuperscript{6}FRC CSC RAS
\\
\href{mailto:vazhentsev@airi.net}{\{vazhentsev, kuzmin, panchenko\}@airi.net}\quad \href{mailto:ekaterina.fadeeva@inf.ethz.ch}{ekaterina.fadeeva@inf.ethz.ch} \\
\href{mailto:artem.shelmanov@mbzuai.ac.ae}{\{rui.xing, preslav.nakov, timothy.baldwin, maxim.panov, artem.shelmanov\}@mbzuai.ac.ae}
}

\begin{document}
\maketitle

\begin{abstract}
  Uncertainty quantification (UQ) has emerged as a promising approach for detecting  hallucinations and low-quality output of Large Language Models (LLMs). However, obtaining proper uncertainty scores is complicated by the conditional dependency between the generation steps of an autoregressive LLM because it is hard to model it explicitly. 
  Here, we propose to learn this dependency from attention-based features. In particular, we train a regression model that leverages LLM attention maps, probabilities on the current generation step, and recurrently computed uncertainty scores from previously generated tokens. To incorporate the recurrent features, we also suggest a two-staged training procedure.
  Our experimental evaluation on ten datasets and three LLMs shows that the proposed method is highly effective for selective generation, achieving substantial improvements over rivaling unsupervised and supervised approaches.\footnote{\url{https://github.com/mbzuai-nlp/llm-tad-uncertainty}}
\end{abstract}

\section{Introduction}

  \begin{figure*}[t!]
    \centering
    \includegraphics[trim={0.cm 0.cm 0.cm 0.cm},clip,width=1.\linewidth]{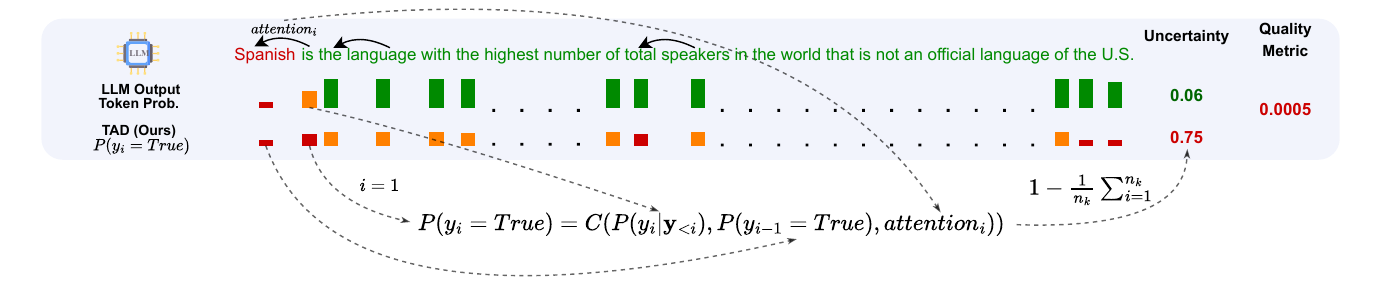}
    \caption{
      An illustration of the proposed method TAD. The figure shows the generated tokens, the uncertainty scores for the generated sequence, and the probabilities assigned by an LLM and by TAD (represented with bars). The output is generated by LLaMa-3.1 8B for the question \textit{What is the language with the highest number of total speakers in the world that is not an official language of the U.S.?} The LLM starts by generating the token \textit{Spanish} that leads to the erroneous answer. The probabilities estimated by the LLM are high for all tokens except for the first one, which makes the uncertainty scores based on raw probabilities misleadingly low. On the contrary, TAD takes into account uncertainty from the previous step using a trainable model $\conf(\cdot)$ based on attention, resulting in a high overall uncertainty for the generated answer.
    }
    \label{fig:tad_vizualization}
  \end{figure*}

  Uncertainty quantification (UQ: \citet{gal2016dropout,fadeeva-etal-2023-lm,baan2023uncertainty,geng2023survey}) is of growing interest in the Natural Language Processing (NLP) community for dealing with hallucinations~\cite{fadeeva2024fact} and low-quality generations~\cite{malinin2020uncertainty} in Large Language Models (LLMs) in an efficient manner. For example, high uncertainty could serve as an indicator that the LLM generation should be discarded as potentially harmful or misleading. This is known in the literature as selective generation~\cite{baan2023uncertainty}.
  There are many approaches for detecting hallucinations and low-quality outputs of LLMs~\cite{ji2023survey,min2023factscore,chen2023hallucination}, but many of them need external knowledge sources or a second LLM. 
  
  Knowledge sources are generally patchy in coverage, while censoring the outputs of a small LLM using a bigger one has a high computational cost and is impractical. 
  We argue that LLMs inherently contain information about the limitations of their own knowledge, and that there should be an efficient way to access this information, which can enable LLM-based applications that are both safe and practical. 

  While a rich body of UQ techniques has been developed for general text classification and regression tasks~\cite{zhang-etal-2019-mitigating,he2020towards,xin-etal-2021-art,wang-etal-2022-uncertainty,vazhentsev-etal-2022-uncertainty,vazhentsev-etal-2023-hybrid,he-etal-2024-uncertainty}, applying UQ to text generation is considerably more challenging. A key difficulty arises from the fact that LLMs generate sequences token by token, making multiple conditionally dependent predictions~\cite{zhang2023enhancing}. 
   Since LLMs generate text by conditioning on previously produced tokens, an early hallucination, whether at the beginning or in the middle of a sequence, can propagate, causing subsequent claims to also be incorrect. Crucially, even if the generation of the first claim was highly uncertain, this uncertainty is not taken into account during the subsequent generation process. This means that although the first error may be recognized due to its high uncertainty, all subsequent errors are overlooked because the generation process conditioned on it proceeds with high confidence. Therefore, effective hallucination detection requires accounting for this dependency and propagating uncertainty across generation steps.

  In this work, we note that the attention between the generated tokens provides information about the conditional dependency between the generation steps. Previously, there have been several attempts to suggest heuristic approaches to model this dependency~\cite{zhang2023enhancing}. 
  
  We argue that the particular algorithmic function would be too difficult to engineer, and thus we propose to learn this dependency from data instead. For this purpose, we generate a training dataset with a target variable, representing the quality score of the generated text according to some ground truth annotation, and train a regression model that leverages LLM attention maps, probabilities on the current generation step, and recurrently computed uncertainty scores from previously generated tokens. To incorporate recurrent features, we suggest a two-staged training procedure where in the second stage, we use scores from the intermediate model obtained in the first training stage. We call the proposed approach \emph{Trainable Attention-based Dependency} (\emph{TAD}). \Cref{fig:tad_vizualization} illustrates the idea of the method on the real output of an LLM. 

  The \textbf{contributions} of this work are as follows:
  \begin{itemize}
    \item  We develop a new data-driven supervised approach to uncertainty quantification that leverages features based on attention maps, probabilities on the current generation step, and recurrently computed uncertainty scores from previously generated tokens.

    \item We show that both attention and recurrent features are essential for achieving high performance in UQ, and a two-step training procedure is necessary to avoid overfitting.

    \item We conduct a comprehensive empirical investigation of selective generation, and show that the proposed approach outperforms existing unsupervised and supervised UQ methods across nine datasets and three LLMs.
  \end{itemize}

\section{Related Work}
  The majority of the methods for UQ for LLMs have been unsupervised, with only a few supervised approaches proposed more recently.

\ecoparagraph{Unsupervised UQ methods.}
  One of the most straightforward approaches to UQ is information-based methods \cite{fomicheva-etal-2020-unsupervised,fadeeva-etal-2023-lm}. These methods rely on probability distributions estimated by LLMs for individual tokens, which are then aggregated into uncertainty scores. A key limitation, however, is that these distributions conflate uncertainties from multiple sources, some of which are irrelevant for assessing generation quality. \citet{fadeeva2024fact} addressed this issue by reducing such sources of uncertainty through probability distribution renormalization.
  
  Another major class of UQ methods is sampling-based~\citep{fomicheva-etal-2020-unsupervised,kuhn2023semantic,lin2023generating,nikitin2024kernel,cheng-vlachos-2024-measuring,zhang-etal-2024-luq,chen2024eigenscore}. The main idea is to generate multiple outputs from an LLM, group them into semantically equivalent clusters, and analyze the diversity of meanings rather than surface-level variations. This approach addresses one of the pivotal problems of UQ for LLMs -- multiple, or a potentially infinite number of correct answers with variable forms.

 Density-based methods quantify uncertainty by measuring distances from a reference distribution in the model’s hidden space. \citet{vazhentsev-etal-2023-efficient} applied the Mahalanobis distance to sequence-to-sequence models for out-of-distribution detection, and \citet{ren2023outofdistribution} extended this approach to selective generation by combining it with probability-based measures.

  \citet{moskvoretskii2025adaptive} addressed the sources of uncertainty in LLMs arising from retrieved context, focusing on determining whether RAG should be used, while \citet{jellybell} examined how to identify the most suitable context for a given query.
  
  \citet{zhang2023enhancing} and~\citet{duan2023shifting} highlighted that not all tokens should contribute to the uncertainty score, proposing heuristics to select the relevant ones. \citet{zhang2023enhancing} modeled the conditional dependencies between the generation steps by penalizing the uncertainty scores based on the uncertainties of the previously generated tokens and the max-pooled attention to previous tokens.

  Overall, most previous work on UQ has not addressed the conditional dependency between the predictions, or has addressed it using heuristics. We argue that the conditional dependency is an important aspect of UQ for text generation tasks, and propose a data-driven approach for dealing with it. We also note that techniques based on sampling multiple answers from LLMs usually introduce prohibitive computational overhead. We argue that for UQ methods to be practical, they should also be computationally efficient.

\paragraph{Supervised UQ methods.}
  Supervised regression-based confidence estimators are well-known for classification problems, primarily from computer vision~\citep{lahlou2022deup,park2024density}. Their key benefit is computational efficiency.
  Some researchers applied this approach to text generation. 
  \citet{lu-etal-2022-learning} proposed training a regression head to predict confidence. They noted that the probability distribution of a language model is poorly calibrated and cannot be used directly to spot low-quality translations. 
  They modified the model architecture and the loss function, restricting this approach to fine-tuning language models only for machine translation and making it unsuitable for general-purpose LLMs.
  In a similar vein, \citet{azaria-mitchell-2023-internal} approached the task of UQ by training a multi-layer perceptron (MLP) on the activations of the internal layers of LLMs to classify true vs.\ false statements. 
  They demonstrated that it outperformed other supervised baselines and few-shot prompting of the LLM itself. However, the reliance on forced decoding limits the real-world applicability to unrestricted generation cases.
  
  Several studies enhanced this method by refining the model architecture and the training procedure. \citet{su-etal-2024-unsupervised} combined the hidden state of the last token with the average hidden state of the sequence, while \citet{ch-wang-etal-2024-androids} introduced a trainable attention layer over token embeddings and used linear regression on top of the MLP's predictions based on embeddings from various layers. \citet{he-etal-2024-llm} proposed to combine multiple deep learning models trained on diverse features extracted from hidden states. \citet{chuang-etal-2024-lookback} suggested training the linear classifier using features derived from attention matrices. \citet{vazhentsev-etal-2025-token} proposed extracting token embeddings from multiple layers of LLMs, computing density-based scores for each token, and training the linear regression on these features. 

  Unlike previous methods, we focus on modeling the conditional dependencies between generation steps using attention in a supervised way. We incorporate recurrently computed uncertainty scores for tokens from previous generation steps, capturing the relationship between the uncertainty of generated tokens. Our method is flexible as it can be applied at different levels: to the entire text, to a sub-sequence, or to individual tokens. Finally, unlike the method proposed by \citet{chuang-etal-2024-lookback}, which relies on feature engineering, ours directly utilizes raw attention weights, which give access to more information.

\vspace{0.4cm}
\section{Problem Background and Key Idea}
  When an LLM generates a token $y_i$, it provides us a conditional probability distribution $p(y_i \mid \yv_{<i}) = p(y_i \mid \xv, \yv_{<i})$, where $\xv$ is an input prompt and $\yv_{<i}$ is a sequence of tokens generated before the token $y_i$. This essentially means that the LLM considers that everything generated so far is correct, which might not be the case. In practice, we would like to somehow propagate the uncertainty from the previous generation steps.
  
  To illustrate the problem, 
  let us assume that only the uncertainty from the previous tokens is propagated to the current generation step, i.e., 
  $p(y_i \mid \yv_{<i}) \simeq p(y_i \mid y_{i-1})$. Let us further consider that we have trained an LLM that generates only tokens that are true (``T'') or false (``F''). The probability of the token $y_i$ being T is given by the conditional probability $p(y_i \mid y_{i-1}) = p(y_i = \ctrue \mid y_{i-1} = \ctrue)$. Assume we already have some tokens $y_1, y_2, \dots, y_n$ and a prompt $\xv$. At each step, the LLM provides us $p(y_1 = \ctrue \mid \xv), p(y_2 = \ctrue \mid y_1 = \ctrue), \dots, p(y_n = \ctrue \mid y_{n-1} = \ctrue)$.

  These probability distributions are conditionally dependent on the previously generated tokens. However, to estimate the correctness of some token $y_i$, we need to obtain an \textit{unconditional probability} $p(y_i) = p(y_i = \ctrue)$. Let us expand $p(y_i = \ctrue)$ according to the law of total probability and express it using conditional probability:
  \begin{align*}
    & p(y_i = \ctrue)
    = p(y_i = \ctrue \mid y_{i-1} = \ctrue) \cdot p(y_{i-1} = \ctrue) \nonumber\\ 
    & + p(y_i = \ctrue \mid y_{i-1}=\cfalse) \cdot \bigl(1 - p(y_{i-1} = \ctrue)\bigr).
  \label{formula:pt}
  \end{align*}
  where $p(y_i = \ctrue \mid y_{i-1} = \ctrue)$ is what the LLM provides during the current generation step in accordance with the specified assumptions, and $p(y_{i-1} = \ctrue)$ is recurrently calculated based on the previous generation step. 
  
  We still do not know the remaining term: $p(y_i = \ctrue \mid y_{i-1} = \cfalse)$. This simplistic example shows that in order to obtain an uncertainty estimate suitable for hallucination detection, we cannot rely solely on the probability distribution provided by the LLM, and we also need to model the conditional dependency of the generation steps. It also makes explicit the need for recurrence in token-level uncertainty computation. 
  
  Attention weights are commonly used in interpretability to illustrate which tokens influenced the model's decision at the current generation step~\cite{zhao2024explainability,tufanov-etal-2024-lm,ferrando-voita-2024-information}.
  However, obtaining an explicit expression that accurately approximates the conditional dependency between the generation steps from attention weights is challenging. The assumptions in our simplistic example do not hold in real LLMs, and thus the predictions on each step depend on multiple previous tokens in a complicated fashion. We suggest to learn this dependency in a supervised way from attention. We propose a feature set for training token-level unconditional confidence scores $\conf$: the attention weights $Att_i$, the token probabilities from the LLM on the current step  $p(y_i \mid \yv_{<i})$, and the recurrently calculated confidence scores on the previous steps $\boldsymbol{\conf}_{<i}$: 
  \begin{equation}
    \conf(y_i) = \conf\bigl(Att_i, p(y_i \mid \yv_{<i}), \boldsymbol{\conf}_{<i}\bigr).
  \end{equation}

\section{Trainable Attention-Based Conditional Dependency}
  We suggest learning unconditional token-level probability estimates from annotated data.

\ecoparagraph{Obtaining targets for learning unconditional probability.}
  In order to obtain the targets $\hat{p}(y_i)$ for the unconditional probability $\conf(y_i)$ for a generated token $y_i \in \yv$ during the training phase, we compute the semantic similarity between the generated answer $\yv$ and the ground truth $\yv^*$: 
  \begin{equation}
  \hat{p}(y_i) = \mathrm{sim}(\yv, \yv^*). 
  \label{eq:gt2}
  \end{equation}
  For generating the targets, we use task-specific similarity measures, such as Accuracy, COMET~\cite{rei-etal-2020-comet}, and AlignScore~\cite{zha-etal-2023-alignscore}.

\ecoparagraph{Generating training data for TAD.}
  We generate the training data for TAD using the original textual training dataset in the following way:
  \begin{enumerate}[topsep=0pt,itemsep=-1ex,partopsep=1ex,parsep=1ex]
    \item For the input prompt $\xv_k$, 
    we use an LLM to generate a text $\yv_k = y_1 y_2 \dots y_{n_k}$ of some length $n_k$ and token probabilities $p(y_i \mid \xv_k, \yv_{<i})$.

    \item For the first generated token $y_1$ in each text, we introduce its unconditional confidence estimate 
    $\hat{p}_k(y_1) = \mathrm{sim}(\yv_k, \yv_k^*)$ according to ~\Cref{eq:gt2}.

    \item For each generated token $y_i$, $i = 2, \dots, n_k$ we construct a feature vector $z_i^k$ that depends on  $N$ preceding tokens.
      The feature vector $z_i^k$ includes: the conditional probabilities $p(y_i \mid \xv_k, \yv_{<i})$ and $p(y_{i-l} \mid \xv_k, \yv_{<i-l})$, for $l = 1, \dots, \min\{N, i-1\}$; the unconditional probabilities' estimates from the previous steps $\hat{p}_k(y_{i-l})$, and the attention weights $a_{i, i-l}$ from the $i$-th token to the $(i-l)$-th token  from all layers and heads. If $N > i - 1$, we pad the feature vector with zeros to ensure they have the same length.
      During the first training stage, $z_i^k$ includes only the conditional probabilities without other features. Consequently, on the first stage of learning, the unconditional probabilities $\hat{p}_k(y_{i-l})$ are not required.
      On the subsequent learning stages it is estimated via the function learned on the previous learning stage. 
  \end{enumerate}
  \vspace{0.1cm}
  
  As a result, for each instance in the training dataset and for each iteration of learning, we generate a sequence of target variables $\hat{p}_k(y_i) = \mathrm{sim}(\yv_k, \yv_k^*)$ and corresponding feature vectors $z_i^k, k = 1, \dots, K, i = 2, \dots, n_k$. 
  We use this dataset to train the model $\conf$. The step-by-step procedure for generating training data is presented in \Cref{alg:tad_training_data} in \Cref{sec:tad_training_data}.

\ecoparagraph{Model for $\conf$ and its training procedure.}
  The training procedure needs the estimates of the unconditional probabilities from the previous steps, and thus
   we perform the procedure twice. In the second stage, we leverage the predictions of the function $\conf$ trained on the first stage as features. This two-step training approach enables us to leverage the conditional dependency of the current step on the previous ones when computing the uncertainty score. Our experiments show that this is essential for achieving good performance.

  We experiment with two regression models for TAD: linear regression (LinReg) and a multi-layer perceptron (MLP). The hyperparameters of the regressors are obtained using five-fold cross-validation on the training dataset. We select the values of the hyperparameters based on the best average PRR, and we use these values to train the regression model on the full training set. The values we used are presented in \Cref{sec:hp_appendix}.

\ecoparagraph{Inference procedure.}
  During inference, we obtain predictions from the LLM, and extract features from the attentions. For the first generated token $y_1$, its unconditional probability is defined as $p(y_1) = p(y_1 \mid \xv_k)$. For each subsequent token, the function $\conf$ computes the predictions recursively, leveraging the attentions, the conditional probabilities, and the unconditional probabilities predicted for the preceding tokens. Finally, in order to compute the sequence-level uncertainty score, we aggregate the token-level confidence scores in the following way:
  \begin{equation}
    U(\yv) = 1 - \frac{1}{n_k}\sum_{i=1}^{n_k} \conf^k(y_i).
  \end{equation}
  We experiment with various other aggregation approaches in the ablation study.

  \begin{table*}[!ht] \resizebox{\textwidth}{!}{
\begin{tabular}{l|c|c|c|c|c|c|c|c|c|c|c|c|c}
\toprule
\multirow{2}{*}{\textbf{UQ Method}} & \textbf{XSum} & \textbf{SamSum} & \textbf{CNN} & \textbf{WMT19} & \textbf{MedQUAD} & \textbf{TruthfulQA} & \textbf{CoQA} & \textbf{SciQ} & \textbf{TriviaQA} & \textbf{MMLU} & \textbf{GSM8k} & \multirow{2}{*}{\multirowcell{\textbf{Mean} \\ \textbf{PRR}}} & \multirow{2}{*}{\multirowcell{\textbf{Mean} \\ \textbf{Rank}}} \\ \cline{2-12}
    & \textbf{AlignScore} & \textbf{AlignScore} & \textbf{AlignScore} & \textbf{Comet} & \textbf{AlignScore} & \textbf{AlignScore} & \textbf{AlignScore} & \textbf{AlignScore} & \textbf{AlignScore} & \textbf{Acc.} & \textbf{Acc.} &  \\\midrule
    
MSP & \large\cellcolor[rgb]{0.953077067017647,0.9210455325882353,0.9030752965411765} .303 & \large\cellcolor[rgb]{0.7365350864941176,0.8055387188078431,0.9853167941313725} .107 & \large\cellcolor[rgb]{0.9783266054882354,0.7990169113588235,0.7386511461764707} .329 & \large\cellcolor[rgb]{0.9849255762058824,0.8479150297647058,0.7906558870392157} .459 & \large\cellcolor[rgb]{0.7179074031529412,0.7853536515764705,0.9758586906411765} .091 & \large\cellcolor[rgb]{0.9802905992117648,0.8812505092627452,0.8339817735509804} .314 & \large\cellcolor[rgb]{0.9781854635254902,0.8875721630666666,0.8432079741549019} .262 & \large\cellcolor[rgb]{0.9841016995,0.8604220499999999,0.8061464956666666} .459 & \large\cellcolor[rgb]{0.9763803588352942,0.8914823988,0.8493228856529411} .527 & \large\cellcolor[rgb]{0.9839369241588236,0.8629234540470588,0.8092446173921568} .535 & \large\cellcolor[rgb]{0.8718771013137254,0.9125626824411764,0.9828988807745098} .310 & \large\cellcolor[rgb]{0.8816813900509803,0.917546110909804,0.9778288382784314} .336 & \large\cellcolor[rgb]{0.915574114,0.9297565972666666,0.9515550793333334} 10.55 \\
\bottomrule
\end{tabular}
}\caption{\label{tab:llama_main_results} PRR$\uparrow$ of UQ methods for the Llama-3.1 8b model. Warmer color indicates better results. The best method is in \textbf{bold}, the second best is \underline{underlined}.}\end{table*}

\section{Experiments and Evaluation}

\subsection{Experimental Setup}
  For the experimental evaluation, we use the LM-Polygraph framework~\cite{fadeeva-etal-2023-lm,vashurin2024benchmarking}. We focus on the task of selective generation~\cite{ren2023outofdistribution}, where generated sequences are rejected if their uncertainty scores exceed some threshold, under the assumption that high uncertainty indicates low quality. Rejecting means that we do not use the model output, and the corresponding queries are processed differently, e.g., they could be further reprocessed manually.

\ecoparagraph{Evaluation measures.}
  Following previous work on UQ in text generation~\cite{malinin2020uncertainty,vashurin2024benchmarking,ielanskyi2025addressing}, we compare UQ methods using the Prediction Rejection Ratio (PRR) metric. PRR quantifies how well an uncertainty score can identify and reject low-quality predictions according to some generation quality measure. The PRR scores are normalized to the range $[0,1]$ by linearly scaling the area under the PR curve between the values obtained with random selection (corresponding to 0) and oracle selection (corresponding to 1). Higher PRR values indicate better quality of the selective generation. Following previous work~\cite{vashurin2024benchmarking}, we compute PRR only up to a rejection threshold of 50\%, as higher thresholds usually are not practical. We use Accuracy, COMET~\cite{rei-etal-2020-comet}, and AlignScore~\cite{zha-etal-2023-alignscore,santilli-etal-2025-revisiting} as generation quality measures. We also use ROC-AUC of detecting incorrect answers as a supplementary metric, as it is also widely adopted in the UQ literature.

\ecoparagraph{Datasets.}
  We consider ten datasets from five text generation tasks: text summarization (TS), machine translation (MT), Question Answering (QA) with long free-form answers, QA with free-form short answers, and multiple-choice QA. A detailed description of all datasets is provided in \Cref{sec:ds_stats}, and the dataset statistics are presented in \Cref{tab:dataset_stat}.

\ecoparagraph{LLMs.}
  We experiment with the following three LLMs: LLaMA-3.1 8b~\cite{llama3}, Gemma-2 9b~\cite{gemma2}, and Qwen-2.5 7b~\cite{qwen2.5}. The values of the inference hyperparameters are given in \Cref{tab:llm_hyperparameters} in \Cref{sec:llm_hp}.

\ecoparagraph{UQ baselines.} 
  The set of unsupervised UQ baselines includes Maximum Sequence Probability (MSP), Mean Token Entropy, and Perplexity~\cite{fomicheva-etal-2020-unsupervised}. They are considered simple yet strong and robust baselines for selective generation across various tasks~\cite{fadeeva-etal-2023-lm}. We also compare our method to unsupervised techniques considered to be state-of-the-art: Lexical Similarity based on ROUGE-L~\cite{fomicheva-etal-2020-unsupervised}, black-box methods (DegMat, Eccentricity, EigValLaplacian: ~\citet{lin2023generating}), Semantic Entropy~\cite{kuhn2023semantic}, hallucination detection with a stronger focus (Focus: \citet{zhang2023enhancing}), Claim-Conditioned Probability (CCP: \citet{fadeeva2024fact}), Shifting Attention to Relevance (SAR: \citet{duan2023shifting}), EigenScore~\cite{chen2024eigenscore}, Semantic Density~\cite{qiu2024semantic}, and Long-text Uncertainty Quantification (LUQ: \citet{zhang-etal-2024-luq}). For sampling-based methods, we use five samples.

  The suite of baselines also includes state-of-the-art supervised methods that use hidden states or attention weights: Factoscope~\cite{he-etal-2024-llm}, SAPLMA~\cite{azaria-mitchell-2023-internal}, MIND~\cite{su-etal-2024-unsupervised}, Sheeps~\cite{ch-wang-etal-2024-androids}, LookBackLens~\cite{chuang-etal-2024-lookback}, and SATRMD~\cite{vazhentsev-etal-2025-token}.

  \begin{table}[t!]\footnotesize  \centering\resizebox{0.48\textwidth}{!}{\begin{tabular}{l|c|c|c|c}
\toprule
\textbf{UQ Method} & \textbf{Llama-3.1 8b} & \textbf{Gemma-2 9b} & \textbf{Qwen-2.5 7b} & \textbf{Mean Rank} \\\midrule

MSP & \cellcolor[rgb]{0.915574114,0.9297565972666666,0.9515550793333334} 10.55 & \cellcolor[rgb]{0.9479408841470589,0.9249530282941176,0.9117495381470588} 10.27 & \cellcolor[rgb]{0.7907430740941177,0.8567252977647059,0.9991571764705882} 12.91 & \cellcolor[rgb]{0.9653342981666666,0.909438499827451,0.8795731953490196} 9.67 \\
Perplexity & \cellcolor[rgb]{0.8863529743019607,0.9194891086196078,0.9746593799568628} 11.36 & \cellcolor[rgb]{0.9276891842254902,0.9318890695490196,0.9382935886568627} 10.91 & \cellcolor[rgb]{0.8933603506784313,0.9224036051843137,0.9699051924745098} 10.45 & \cellcolor[rgb]{0.9813541391647058,0.8767786732784314,0.8278006056137255} 8.33 \\
Mean Token Entropy & \cellcolor[rgb]{0.8792694128431373,0.9163932970294117,0.9792039273627451} 11.55 & \cellcolor[rgb]{0.9418435698882353,0.9280538589764706,0.9201288350882354} 10.45 & \cellcolor[rgb]{0.888688766427451,0.9204606074745099,0.9730746507960784} 10.55 & \cellcolor[rgb]{0.9790880158705882,0.8856170452000001,0.8401505184058824} 8.67 \\
CCP & \cellcolor[rgb]{0.7961779297490197,0.861396014627451,0.9997169374117647} 13.64 & \cellcolor[rgb]{0.8694129974705882,0.9112858109117647,0.9841305319117647} 12.64 & \cellcolor[rgb]{0.7285232392627451,0.797002774964706,0.9815146148450979} 14.36 & \cellcolor[rgb]{0.7961779297490197,0.861396014627451,0.9997169374117647} 15.67 \\
Simple Focus & \cellcolor[rgb]{0.9176723556764705,0.9302569986470588,0.9494852049705882} 10.45 & \cellcolor[rgb]{0.9678868848333333,0.9061183506196079,0.873578236792157} 9.45 & \cellcolor[rgb]{0.8492270432274509,0.8997249420568627,0.9922887283509805} 11.55 & \cellcolor[rgb]{0.9840526685,0.8342375980588236,0.7752431104705884} 7.00 \\
Focus & \cellcolor[rgb]{0.8415278407803921,0.8950213134450979,0.9948842140921569} 12.55 & \cellcolor[rgb]{0.9497716033000001,0.923750118,0.9088945372} 10.18 & \cellcolor[rgb]{0.7205613621803921,0.7882659324235295,0.9772726716921569} 14.55 & \cellcolor[rgb]{0.8816813900509803,0.917546110909804,0.9778288382784314} 13.00 \\
Lexical Similarity Rouge-L & \cellcolor[rgb]{0.6892991246352941,0.7519281085960785,0.9568458054235294} 16.18 & \cellcolor[rgb]{0.7825907906117646,0.8497192224705882,0.9983175350588236} 14.82 & \cellcolor[rgb]{0.7446232039529412,0.8137680277960784,0.9884477548843138} 14.00 & \cellcolor[rgb]{0.7608481404156863,0.8297993031764705,0.9938680116235294} 16.67 \\
EigenScore & \cellcolor[rgb]{0.61490285,0.649358983,0.8768415764999999} 18.18 & \cellcolor[rgb]{0.61490285,0.649358983,0.8768415764999999} 19.18 & \cellcolor[rgb]{0.6545301595333333,0.7067491361333333,0.9250638169333334} 16.18 & \cellcolor[rgb]{0.61490285,0.649358983,0.8768415764999999} 21.33 \\
EVL NLI Score entail. & \cellcolor[rgb]{0.7554121621254901,0.8246983074117646,0.9925393881882354} 14.55 & \cellcolor[rgb]{0.7771559349568626,0.8450485056078432,0.9977577741176471} 14.91 & \cellcolor[rgb]{0.7825907906117646,0.8497192224705882,0.9983175350588236} 13.09 & \cellcolor[rgb]{0.7717200969960785,0.8400012947058824,0.9965252584941177} 16.33 \\
Ecc. NLI Score entail. & \cellcolor[rgb]{0.6643022236588235,0.7198559149882353,0.9347936312705882} 16.82 & \cellcolor[rgb]{0.6766845796941177,0.736034329145098,0.9462852021294117} 17.45 & \cellcolor[rgb]{0.6842533066588236,0.7455705968156863,0.9526215641058824} 15.45 & \cellcolor[rgb]{0.6741616707058824,0.7328555732549019,0.9441730814705882} 19.33 \\
DegMat NLI Score entail. & \cellcolor[rgb]{0.8230564053823529,0.8822182482647059,0.9984342312529412} 13.00 & \cellcolor[rgb]{0.8204138911862745,0.8803757532058823,0.9989228874411764} 13.91 & \cellcolor[rgb]{0.815044265017647,0.8762581198529411,0.9992540061705882} 12.36 & \cellcolor[rgb]{0.8517934440431373,0.9012928182607842,0.9914235664372548} 14.00 \\
Semantic Entropy & \cellcolor[rgb]{0.6594161915960784,0.7133025255607843,0.9299287241019608} 16.91 & \cellcolor[rgb]{0.707400451427451,0.7734367635137256,0.9695437661686275} 16.64 & \cellcolor[rgb]{0.6171885397254901,0.652770865164706,0.8798397637941177} 17.18 & \cellcolor[rgb]{0.6520871435019608,0.7034724414196079,0.9226313633490196} 20.00 \\
SAR & \cellcolor[rgb]{0.8440942415960784,0.8965891896490196,0.9940190521784313} 12.45 & \cellcolor[rgb]{0.9024823794117647,0.9258330802784314,0.9630825372156863} 11.73 & \cellcolor[rgb]{0.8256989195784313,0.8840607433235294,0.9979455750647059} 12.09 & \cellcolor[rgb]{0.9276891842254902,0.9318890695490196,0.9382935886568627} 11.33 \\
LUQ & \cellcolor[rgb]{0.7988883877470588,0.8636648624411765,0.9998883658882353} 13.55 & \cellcolor[rgb]{0.8230564053823529,0.8822182482647059,0.9984342312529412} 13.82 & \cellcolor[rgb]{0.8230564053823529,0.8822182482647059,0.9984342312529412} 12.18 & \cellcolor[rgb]{0.8718771013137254,0.9125626824411764,0.9828988807745098} 13.33 \\
Semantic Density & \cellcolor[rgb]{0.8694129974705882,0.9112858109117647,0.9841305319117647} 11.82 & \cellcolor[rgb]{0.9046643351764705,0.9264869997764706,0.9611612942352941} 11.64 & \cellcolor[rgb]{0.8204138911862745,0.8803757532058823,0.9989228874411764} 12.27 & \cellcolor[rgb]{0.931695915645098,0.9325418979098039,0.9338169421313726} 11.17 \\
Factoscope & \cellcolor[rgb]{0.6402749751176471,0.6867115845411764,0.909005371654902} 17.45 & \cellcolor[rgb]{0.6194742294509804,0.6561827473294117,0.8828379510882353} 19.00 & \cellcolor[rgb]{0.61490285,0.649358983,0.8768415764999999} 17.27 & \cellcolor[rgb]{0.61490285,0.649358983,0.8768415764999999} 21.33 \\
SAPLMA & \cellcolor[rgb]{0.8694129974705882,0.9112858109117647,0.9841305319117647} 11.82 & \cellcolor[rgb]{0.9691631781666666,0.9044582760156863,0.8705807575137254} 9.36 & \cellcolor[rgb]{0.7392311256470588,0.8082818218039216,0.9863604477156862} 14.09 & \cellcolor[rgb]{0.9418435698882353,0.9280538589764706,0.9201288350882354} 10.83 \\
MIND & \cellcolor[rgb]{0.9830083599196078,0.8230648707941176,0.7629451741294118} 6.36 & \cellcolor[rgb]{0.9846442845,0.8424908735411765,0.7844876631294118} 7.55 & \cellcolor[rgb]{0.9781854635254902,0.8875721630666666,0.8432079741549019} 7.36 & \cellcolor[rgb]{0.9591408362921569,0.742086736090196,0.6888969625352941} 4.67 \\
Sheeps & \cellcolor[rgb]{0.9053078374137256,0.6343985308588236,0.6177138057666667} \underline{3.09} & \cellcolor[rgb]{0.9837721488176471,0.8654248580941176,0.812342739117647} 8.09 & \cellcolor[rgb]{0.9708639649117647,0.7732067385098039,0.7148535351862745} 4.73 & \cellcolor[rgb]{0.9283579338254901,0.6773519275058824,0.6427436489686275} 3.33 \\
LookBackLens & \cellcolor[rgb]{0.9513294646843138,0.7239693735803922,0.6748606867705882} 4.45 & \cellcolor[rgb]{0.9385746672999999,0.6973227293333333,0.6558619128666667} 4.64 & \cellcolor[rgb]{0.9423217193470588,0.7050085489411765,0.6612532746235295} \underline{3.55} & \cellcolor[rgb]{0.9150932609745098,0.6523663817764705,0.6274457140333334} \underline{2.83} \\
SATRMD & \cellcolor[rgb]{0.9513294646843138,0.7239693735803922,0.6748606867705882} 4.45 & \cellcolor[rgb]{0.9175136022176471,0.6568221562117647,0.6298915758705882} \underline{4.00} & \cellcolor[rgb]{0.9813503623666666,0.8141088755,0.7538180498} 5.55 & \cellcolor[rgb]{0.9240202011823531,0.6691400189882354,0.6376028197294119} 3.17 \\\midrule
TAD & \cellcolor[rgb]{0.852836579,0.50777808,0.575116406} \textbf{1.82} & \cellcolor[rgb]{0.852836579,0.50777808,0.575116406} \textbf{2.36} & \cellcolor[rgb]{0.852836579,0.50777808,0.575116406} \textbf{1.27} & \cellcolor[rgb]{0.852836579,0.50777808,0.575116406} \textbf{1.00} \\

\bottomrule
\end{tabular}}\caption{\label{tab:average_results} Mean ranks of UQ methods aggregated over all datasets for each LLM separately (the lower the better). The column \emph{Mean Rank} corresponds to the mean rank of the ranks across all LLMs. %Warmer colors indicate better results. 
The best method is in \textbf{bold}, the second best is \underline{underlined}.}
\end{table}

\subsection{Main Results}

\paragraph{Fine-grained comparison to the baselines.}
  \Cref{tab:llama_main_results,tab:gemma_main_results,tab:qwen_main_results} in \Cref{sec:appendix_main_results} present the results for LLaMa-3.1 8b, Gemma-2 9b, and Qwen-2.5 7b, respectively. 
  We can see that across all summarization and translation datasets, both LookBackLens and TAD outperform the current state-of-the-art methods by a sizable margin. 
  
  For Llama, LookBackLens achieves slightly better results than TAD on the WMT19 dataset, but TAD confidently outperforms LookBackLens on all summarization datasets. For Qwen, TAD consistently achieves the best results on all summarization and translation datasets, while LookBackLens achieves the second-best results.

  For QA involving long answers (e.g., MedQUAD, TruthfulQA), TAD consistently demonstrates substantial improvements over the baselines across all considered models. For example, in the experiment with LLaMA-3.1 8b on MedQUAD, TAD outperforms the second-best baseline, LookBackLens, by 0.077 of PRR. On the TruthfulQA dataset, TAD achieves an improvement of 0.045 in PRR over the second-best baseline with Gemma.
  
  On the GSM8k dataset, TAD consistently demonstrates strong performance and outperforms unsupervised methods, although it performs slightly worse than the Sheeps method.
  
  For QA tasks with short answers (CoQA, SciQ, and TriviaQA), TAD generally yields substantial improvements over baseline methods in most cases. The only exception is the SciQ dataset, where LookBackLens is marginally better for LLaMA-3.1 8b and Gemma-2 9b. On TriviaQA, when using LLaMA-3.1 8b, TAD outperforms sampling-based methods, while other supervised methods fall behind simple baselines by a margin. 

  Finally, for MMLU, TAD also notably outperforms state-of-the-art methods for both Gemma-2 9b and Qwen-2.5 7b. However, for LLaMA-3.1 8b, TAD slightly falls behind MIND.

  Summarizing, our findings indicate that certain UQ methods, such as LookBackLens, SATRMD, and Sheeps, achieve top performance in specific experimental settings. However, TAD demonstrates the most consistent and robust performance across all eleven tasks, never ranking below the state-of-the-art unsupervised methods.
  Other supervised methods occasionally underperform, sometimes even falling below simple baselines such as MSP.
  Similar patterns are observed in the ROC AUC results reported in \Cref{tab:llama_roc_auc_results,tab:gemma_roc_auc_results,tab:qwen_roc_auc_results} (see \Cref{sec:appendix_roc_auc}). 

\ecoparagraph{Aggregated results.}
  \Cref{tab:average_results} shows the mean rank aggregated over all datasets for each model separately (lower ranks are better). The column \emph{Mean Rank} shows the mean rank across all models. 
  
\begin{figure}[t!]
    \centering
    \includegraphics[trim={0.3cm 0.cm 0.cm 0.cm},clip,width=1.0\linewidth]{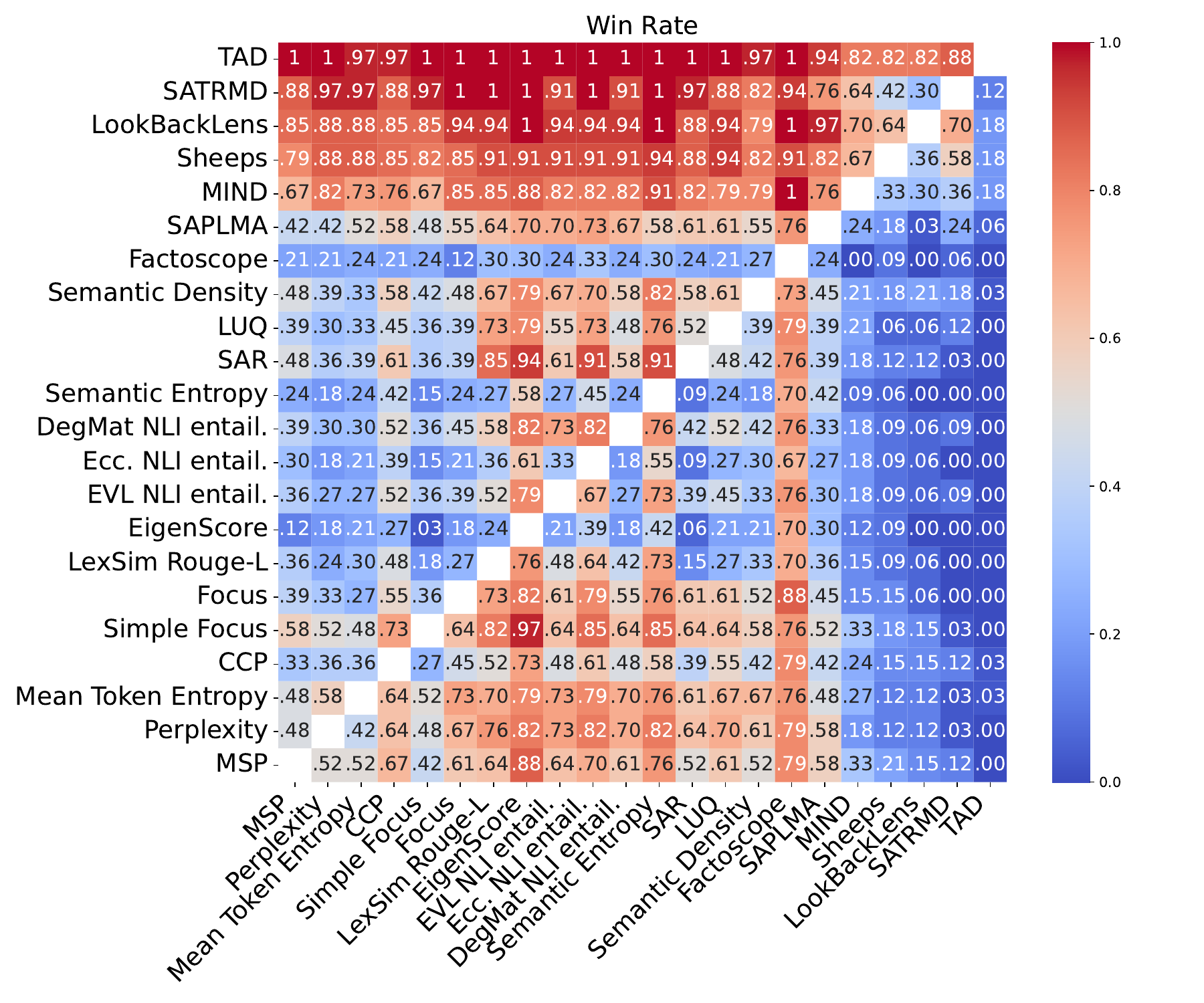}
    \caption{
      Summary of 33 experimental setups with various models and datasets. Each cell in the diagram presents a fraction of experiments where a method from a row outperforms a method from a column. Warmer colors indicate better results.
    }
    \label{fig:win_rates}
  \end{figure}

  \Cref{fig:win_rates} additionally summarizes all experimental setups. Each cell presents a win rate for a method from a row compared to a method from a column. The aggregated results emphasize the significance of the performance improvements of the proposed method. Despite some baselines showing good results in particular cases, they are usually quite unstable, resulting in poor overall ranking. TAD is more robust across multiple tasks and LLMs, which makes it a better choice overall. 

\ecoparagraph{Generalization to out-of-domain datasets.}
  \Cref{tab:llama_generalization_qa_results} compares the results of the supervised methods trained on all QA datasets except for one that represents the out-of-domain dataset for testing. Additionally, \Cref{tab:llama_generalization_mt_ats_results} in \Cref{sec:appendix_generalization} presents the results when these methods are trained on all QA datasets and tested on the out-of-distribution tasks: summarization and translation. These settings evaluate the out-of-domain generalization capabilities of the supervised techniques for both new domains and new tasks. 
  
  The results show that all supervised methods substantially degrade compared to their in-domain performance and, in many cases, underperform the simple MSP baseline. Nevertheless, TAD demonstrates strong out-of-domain performance on the QA datasets, outperforming MSP by 0.015 PRR on average. However, all supervised methods perform significantly worse than MSP on the OOD tasks, summarization and translation, underscoring their limited adaptability to unseen tasks.

  \begin{table}[t] \centering\resizebox{0.48\textwidth}{!}{\begin{tabular}{l|c|c|c|c|c|c}
\toprule
\multirow{2}{*}{\textbf{UQ Method}} & \textbf{CoQA} & \textbf{SciQ} & \textbf{TriviaQA} & \textbf{MMLU} & \textbf{GSM8k} & \multirow{2}{*}{\multirowcell{\textbf{Mean} \\ \textbf{PRR}}} \\ \cline{2-6}
    & \textbf{AlignScore} & \textbf{AlignScore} & \textbf{AlignScore} & \textbf{Acc.} & \textbf{Acc.} &  \\
\midrule

MSP & \large\cellcolor[rgb]{0.9836582578333333,0.8287354144039216,0.7690800753647059} .262 & \large\cellcolor[rgb]{0.9441952453705882,0.708851458745098,0.6639489555019608} \underline{.459} & \large\cellcolor[rgb]{0.8979687697431373,0.6209226426705883,0.6104148745666667} .527 & \large\cellcolor[rgb]{0.852836579,0.50777808,0.575116406} \textbf{.535} & \large\cellcolor[rgb]{0.8763519705509804,0.5787878133490196,0.5921289546450981} .310 & \large\cellcolor[rgb]{0.8763519705509804,0.5787878133490196,0.5921289546450981} \underline{.419} \\
SAR & \large\cellcolor[rgb]{0.9659156483,0.7595427616,0.7032398043} \underline{.297} & \large\cellcolor[rgb]{0.9622044108411765,0.7492947789176471,0.6945295061352941} .439 & \large\cellcolor[rgb]{0.8734190061058824,0.5700105097411765,0.5899980484784314} .552 & \large\cellcolor[rgb]{0.9691631781666666,0.9044582760156863,0.8705807575137254} .275 & \large\cellcolor[rgb]{0.8587172724588235,0.5255587742117647,0.5793683038509804} \underline{.320} & \large\cellcolor[rgb]{0.9348276152529411,0.6896369097254902,0.6504705511098039} .377 \\
Semantic Density & \large\cellcolor[rgb]{0.852836579,0.50777808,0.575116406} \textbf{.380} & \large\cellcolor[rgb]{0.9560162876490197,0.7348397910862745,0.6832824522294118} .448 & \large\cellcolor[rgb]{0.852836579,0.50777808,0.575116406} \textbf{.571} & \large\cellcolor[rgb]{0.9438760079745099,0.9270202487490196,0.9173357361078431} .237 & \large\cellcolor[rgb]{0.9849255762058824,0.8479150297647058,0.7906558870392157} .197 & \large\cellcolor[rgb]{0.9441952453705882,0.708851458745098,0.6639489555019608} .366 \\\midrule
Factoscope & \large\cellcolor[rgb]{0.6817303976705882,0.7423918409254902,0.9505094434470589} .016 & \large\cellcolor[rgb]{0.61490285,0.649358983,0.8768415764999999} .055 & \large\cellcolor[rgb]{0.876805309,0.9151164254999999,0.9804355785000001} .161 & \large\cellcolor[rgb]{0.7880256462666666,0.8543899393333334,0.9988772960000001} .078 & \large\cellcolor[rgb]{0.8204138911862745,0.8803757532058823,0.9989228874411764} .049 & \large\cellcolor[rgb]{0.7446232039529412,0.8137680277960784,0.9884477548843138} .072 \\
SAPLMA & \large\cellcolor[rgb]{0.61490285,0.649358983,0.8768415764999999} -.030 & \large\cellcolor[rgb]{0.8123516188058824,0.8741592436176471,0.9993597327901961} .199 & \large\cellcolor[rgb]{0.61490285,0.649358983,0.8768415764999999} -.112 & \large\cellcolor[rgb]{0.61490285,0.649358983,0.8768415764999999} -.089 & \large\cellcolor[rgb]{0.61490285,0.649358983,0.8768415764999999} -.077 & \large\cellcolor[rgb]{0.61490285,0.649358983,0.8768415764999999} -.022 \\
MIND & \large\cellcolor[rgb]{0.7285232392627451,0.797002774964706,0.9815146148450979} .044 & \large\cellcolor[rgb]{0.7446232039529412,0.8137680277960784,0.9884477548843138} .153 & \large\cellcolor[rgb]{0.9377786937156862,0.9301210794313726,0.9257150330490196} .237 & \large\cellcolor[rgb]{0.9547297988764706,0.919693239882353,0.9001656762117647} .252 & \large\cellcolor[rgb]{0.9747269947588235,0.7861939559392157,0.7267214884509805} .230 & \large\cellcolor[rgb]{0.9068462909411765,0.9271409192745098,0.959240051254902} .183 \\
Sheeps & \large\cellcolor[rgb]{0.8096589725941177,0.8720603673823529,0.9994654594098039} .092 & \large\cellcolor[rgb]{0.9729271893941176,0.7797824782294118,0.7207566595882353} .422 & \large\cellcolor[rgb]{0.9727701494549019,0.8993028702666667,0.8615527086490196} .295 & \large\cellcolor[rgb]{0.9575785619705882,0.7384632635882353,0.686089707382353} .425 & \large\cellcolor[rgb]{0.852836579,0.50777808,0.575116406} \textbf{.323} & \large\cellcolor[rgb]{0.9813503623666666,0.8141088755,0.7538180498} .312 \\
LookBackLens & \large\cellcolor[rgb]{0.7880256462666666,0.8543899393333334,0.9988772960000001} .079 & \large\cellcolor[rgb]{0.9834812190705882,0.867835001309804,0.8154382697392157} .365 & \large\cellcolor[rgb]{0.9754778064901961,0.8934375166666666,0.8523803414019608} .304 & \large\cellcolor[rgb]{0.9591408362921569,0.742086736090196,0.6888969625352941} .422 & \large\cellcolor[rgb]{0.9754778064901961,0.8934375166666666,0.8523803414019608} .166 & \large\cellcolor[rgb]{0.9808223691882353,0.8790145912705882,0.830891189582353} .267 \\
SATRMD & \large\cellcolor[rgb]{0.9844312501823529,0.8554192419058824,0.7999502522156863} .247 & \large\cellcolor[rgb]{0.9781854635254902,0.8875721630666666,0.8432079741549019} .349 & \large\cellcolor[rgb]{0.9460687713941176,0.7126943685490196,0.6666446363803922} .469 & \large\cellcolor[rgb]{0.9176723556764705,0.9302569986470588,0.9494852049705882} .205 & \large\cellcolor[rgb]{0.8763519705509804,0.5787878133490196,0.5921289546450981} .311 & \large\cellcolor[rgb]{0.9802450306647059,0.8081382119705882,0.7477333002470589} .316 \\\midrule
TAD & \large\cellcolor[rgb]{0.9756268974411765,0.7893996947941176,0.729703902882353} .283 & \large\cellcolor[rgb]{0.852836579,0.50777808,0.575116406} \textbf{.529} & \large\cellcolor[rgb]{0.8587172724588235,0.5255587742117647,0.5793683038509804} \underline{.565} & \large\cellcolor[rgb]{0.8790560841823529,0.5840607685176471,0.5944135352882353} \underline{.512} & \large\cellcolor[rgb]{0.9261890675039215,0.6732459732470588,0.6401732343490196} .278 & \large\cellcolor[rgb]{0.852836579,0.50777808,0.575116406} \textbf{.434} \\

\bottomrule
\end{tabular}
}\caption{\label{tab:llama_generalization_qa_results} PRR$\uparrow$ for Llama-3.1 8b model for various QA tasks for the considered supervised sequence-level methods trained on the general QA dataset. Unsupervised methods are not included as their performance is not dependent of the training data. Warmer colors indicate better results. The best method is in \textbf{bold}, and the second best one is \underline{underlined}.}\end{table}
  
  These findings indicate that previous supervised UQ methods are generally effective only for in-domain selective generation. However, the TAD method demonstrates the ability to achieve generalization to unseen domains within similar tasks. More details about these experiments are presented in \Cref{sec:appendix_generalization}.

\subsection{Ablation Studies}

  \begin{figure*}[t!]
    \centering
    \begin{subfigure}[b]{0.52\textwidth}
        \includegraphics[width=\textwidth]{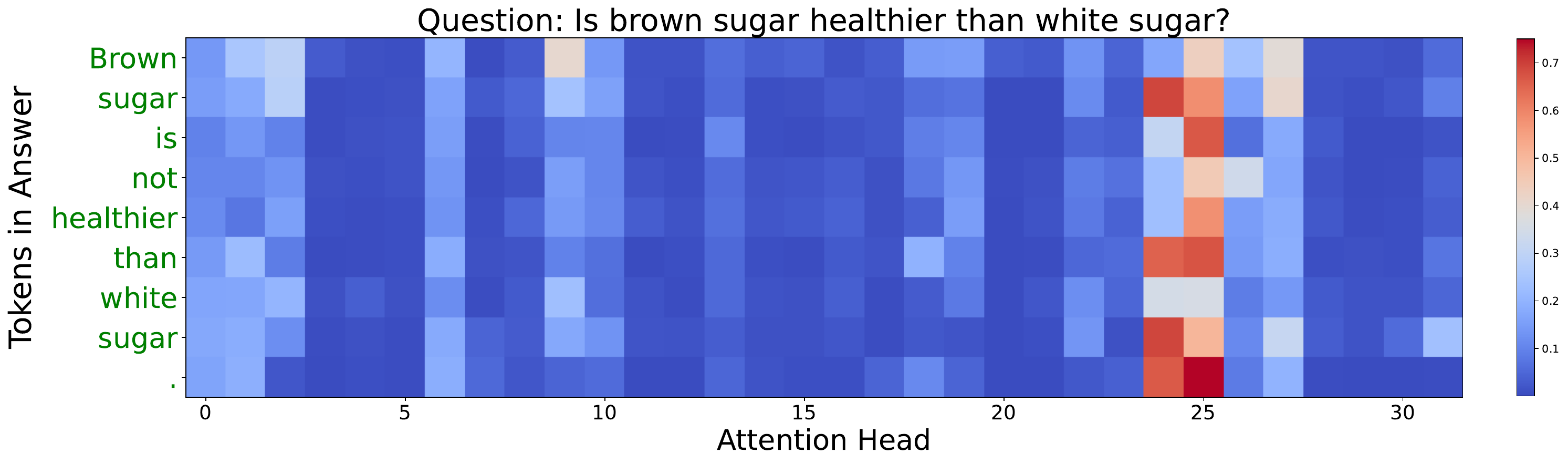}
        \caption{Correct answer}
        \label{fig:attention_correct}
    \end{subfigure}
    \hfill
    \begin{subfigure}[b]{0.44\textwidth}
        \includegraphics[width=\textwidth]{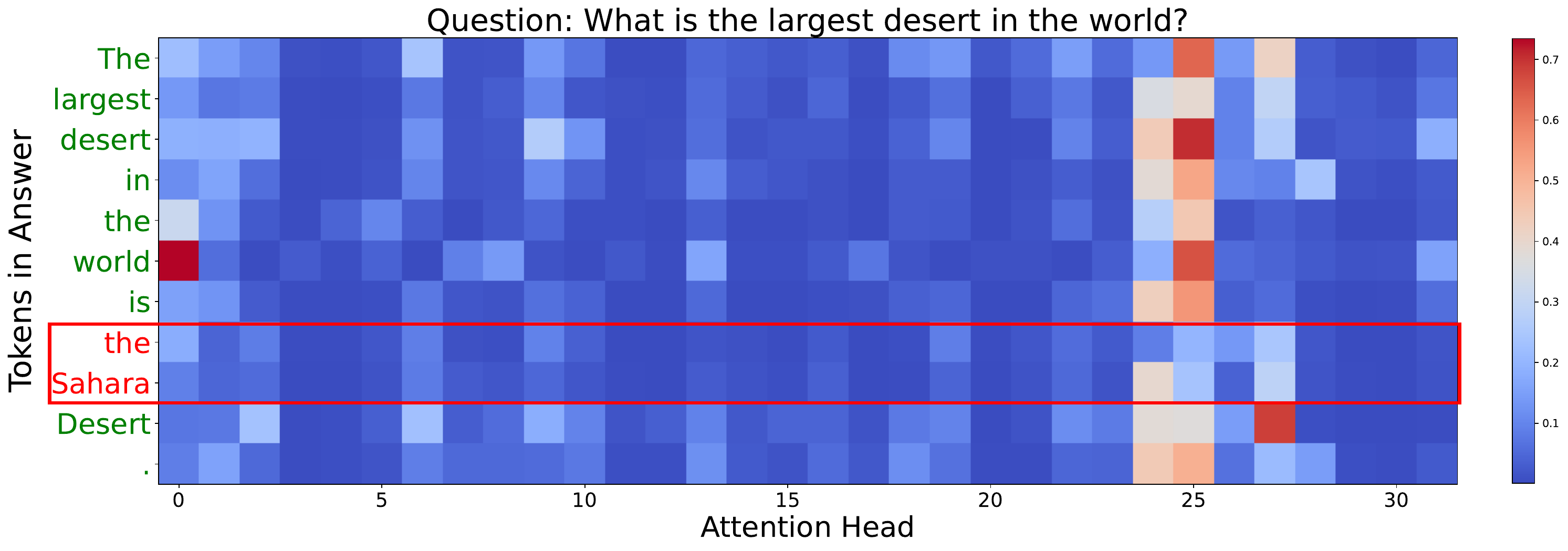}
        \caption{Incorrect answer}
        \label{fig:attention_incorrect}
    \end{subfigure}
    \caption{
    Comparison of the attention weights of Llama-3.1 8B to the last preceding token for each generated token for correct and incorrect answers to input questions from the TruthfulQA dataset. The $y$-axis shows the generated tokens, and the $x$-axis represents the attention heads in the 30th layer. Warmer colors indicate higher attention values.
    In the incorrect answer (Figure~\subref{fig:attention_incorrect}), the model hallucinates the factually incorrect tokens \textit{The Sahara} (the correct answer is Antarctica). Notably, while the 25th attention head consistently assigns high attention to the preceding token in both outputs, this attention noticeably drops for the hallucinated tokens \textit{The Sahara}. This decrease in attention could serve as a valuable signal for a hallucination detector in the TAD method.}
    \label{fig:attention_vizualization}
  \end{figure*}

\ecoparagraph{Comparison of features.}
  \Cref{tab:llama_ablation_results} in \Cref{sec:appendix_ablation} presents the ablation experiment with different features for the TAD regression model. For \emph{TAD (probs.)}, we only use probabilities along with predictions from the preceding tokens $p(y_{i-k} = \ctrue)$ for $k = 1, \dots, N$. For \emph{TAD (attention)}, we use attention weights on the $N$ preceding tokens without probabilities. The results show that \emph{TAD (probs.)} provides meaningful but usually lower performance. \emph{TAD (attention)} demonstrates substantial improvements, underscoring the importance of using the attentions in the TAD method. Finally, \emph{TAD (attention+probs.)}, which combines all of attention weights, probabilities, and uncertainty scores from previous steps, achieves slight but consistent performance gains. This indicates the benefit of recurrence during the computation of uncertainty scores.

\ecoparagraph{Impact of the token-level training procedure.}
  \Cref{tab:llama_taq_seq} in \Cref{sec:appendix_ablation} presents an ablation study comparing different training procedures for the regression model in TAD. We compare the original TAD against \emph{TAD (Sequence-level)}, which uses a two-layer MLP with averaging of the hidden features between layers, followed by a linear layer for direct sequence-level uncertainty prediction. 
  
  The results demonstrate that while \emph{TAD (Sequence-level)} is competitive: the original TAD surpasses it by 0.027 PRR on average, with the largest improvement of 0.124 PRR on MedQUAD. This highlights the effectiveness of the token-level training procedure with recurrent features in TAD.
  
\ecoparagraph{Impact of the two-step training procedure.}
  \Cref{tab:llama_steps_results} in \Cref{sec:appendix_ablation} presents the ablation experiment comparing one-step vs. two-step training procedures for the TAD method. The results show that the two-step procedure is essential for training a well-performing recurrent model.

\ecoparagraph{Regression models and aggregation approaches.}
  Detailed results with various regression models and aggregation approaches are presented in \Cref{tab:llama_models_results}. The optimal values of the hyperparameters of TAD for all experimental setups are presented in \Cref{tab:llama_tad_hp,tab:gemma_tad_hp,tab:qwen_tad_hp} in \Cref{sec:hp_appendix} for LLaMA-3.1 8b, Gemma-2 9b, and Qwen-2.5 7b, respectively. 

  We compared two strategies for aggregating the token-level TAD scores: (\emph{i})~the mean of the scores; and (\emph{ii})~the sum of the log scores inspired by perplexity. For the majority of the settings, the mean of the probabilities yielded the best results. However, for QA with short answers, the sum of the log probabilities performed slightly better. Yet, the difference between MLP and LinReg is minimal. On average, TAD with LinReg outperforms TAD with MLP by 0.013 PRR. Therefore, for simplicity, we use LinReg as a regression method for TAD. 

\ecoparagraph{Impact of the number of previous tokens.}
  \Cref{tab:llama_tokens_results} in \Cref{sec:appendix_ablation} presents results with different numbers of preceding tokens used in TAD. It shows that using ten preceding tokens generally yields better performance compared to using only 1--2 tokens across all datasets, except for XSum.

\ecoparagraph{Impact of the attention layers.}
  \Cref{fig:attention_layers} in \Cref{sec:appendix_ablation} presents the normalized average weights of linear regression for different attention layers in the TAD method. We can see similar patterns across various tasks, revealing that the most important layers are typically the middle ones, which is consistent with observations in previous work~\cite{azaria-mitchell-2023-internal,chen2024eigenscore}. We further note that for the majority of the tasks, the first and last attention layers play a crucial role.

\ecoparagraph{Replacing attention weights with interpretability features.}
  \Cref{tab:llama_lig_res_reordered} in \Cref{sec:appendix_lig} shows the results, where we investigate interpretability features from Layer Integrated Gradients (LIG: ~\citet{lig}) as a measure of conditional dependency between generation steps. We compare the original TAD method with two variants: \emph{TAD (LIG)}, which replaces attention weights with LIG features, and \emph{TAD (MIX)}, which concatenates LIG features with the raw attention weights. LIG features perform comparably to attention, but their inclusion does not enhance TAD performance.

\subsection{Analysis of Attention Maps}
  To gain deeper insight into the mechanisms driving the strong performance of the TAD method, we analyzed the attention maps used as features in the hallucination detector for Llama-3.1 8B. \Cref{fig:attention_vizualization} shows the attention weights assigned to the immediately preceding token of each generated token, comparing correct and incorrect answers to questions from the TruthfulQA dataset. We focus exclusively on attention weights from the current token to its immediate predecessor. We chose this focus because~\Cref{tab:llama_tokens_results} in~\Cref{sec:appendix_ablation} shows that relying solely on attention to the previous token still achieves strong performance. 

  This analysis reveals several key patterns in the attention weights that TAD may leverage. First, there is a small number of attention heads that usually pay high attention to the previous token. In our example, the highest average attention across all generated tokens was expressed in the 30th layer by the 25th head. Second, during correct generation, all tokens assign high attention to previous tokens in these heads, whereas in hallucinated outputs, this attention becomes blurred.
  For instance, in \Cref{fig:attention_correct}, all tokens are correct, resulting in consistently high attention from the 25th head. In contrast, in \Cref{fig:attention_incorrect}, the LLM produces a hallucination \emph{The Sahara} (the correct answer is \emph{Antarctica}), and the attention to this token drops noticeably. This decrease in attention assumably provides a valuable signal to the TAD method.

\subsection{Computational Efficiency}

  \begin{table}[t]\resizebox{\linewidth}{!}{ \centering\begin{tabular}{l|c|c}
\toprule
\textbf{UQ Method} & \textbf{\multirowcell{Runtime \\ per batch}} & \textbf{\multirowcell{Overhead}} \\
\midrule
MSP & $1.30$\tiny{$\pm$$0.62$} & - \\
\midrule
DegMat NLI Score Entail. & $6.86$\tiny{$\pm$$2.28$} & 430 \%\\
Lexical Similarity ROUGE-L & $6.72$\tiny{$\pm$$2.24$} & 420\% \\
Semantic Entropy & $6.86$\tiny{$\pm$$2.28$} & 430\% \\
SAR & $8.83$\tiny{$\pm$$2.94$} & 580\% \\
\midrule
Factoscope & $3.30$\tiny{$\pm$$2.13$} & 150\% \\
SAPLMA & $1.30$\tiny{$\pm$$0.62$} & \textbf{0.06\%} \\
MIND & $1.30$\tiny{$\pm$$0.62$} & 0.10\% \\
Sheeps & $1.50$\tiny{$\pm$$0.97$} & 15\% \\
LookBackLens & $1.30$\tiny{$\pm$$0.62$} & \underline{0.08\%} \\
SATRMD & $1.39$\tiny{$\pm$$0.67$} & 8\% \\
\midrule
TAD & $1.37$\tiny{$\pm$$0.68$} & 5\% \\

\bottomrule
\end{tabular}
}
\caption{\label{tab:llama_comp_efficiency} Evaluation of the inference runtime of UQ methods measured on all test instances from all datasets with predictions from Llama-3.1 8b. The best results are in \textbf{bold}, and the second best results are \underline{underlined}.}
\end{table}

  To demonstrate the computational efficiency of TAD, we compare its runtime to other UQ methods. We use a single 80GB H100 GPU, as detailed in~\Cref{tab:llama_main_results}. The inference is implemented as a single-batch model call for all tokens in the output.

  \Cref{tab:llama_comp_efficiency} presents the average runtime per text instance for each UQ method, along with the percentage overhead over standard LLM inference with MSP. We can see that many state-of-the-art UQ methods (DegMat, Lexical Similarity, Semantic Entropy, and SAR) introduce huge computational overhead (400--600\%) as they need to perform sampling from the LLM multiple times. 
  
  In contrast, all supervised methods we experimented with introduce minimal overhead.  In particular, TAD introduces only 5\% overhead, which makes it a highly practical and efficient choice for uncertainty quantification.

\section{Conclusion and Future Work}
  We introduced a new supervised uncertainty quantification method that learns conditional dependencies between predictions across multiple generation steps. The approach leverages attention to construct features capturing these functional dependencies and recurrently applies them to adjust the uncertainty of subsequent generation steps. This helps to improve results in selective generation tasks, especially when the LLM output is long. Our experimental study shows that TAD usually outperforms other state-of-the-art UQ methods (such as SAR) resulting in the best overall performance across three LLMs and nine datasets. Contrary to other supervised methods, TAD also shows cross-domain generalization. Our method requires only minimal computational overhead due to the simplicity of the underlying linear regression model, making it a practical choice for LLM-based applications. 
  
  In future work, we plan to apply the suggested method to UQ of retrieval-augmented LLMs. In particular, TAD could potentially be used to assess the credibility of a retrieved piece of textual evidence.

\section*{Limitations}
  The proposed approach is supervised and thus benefits from task-specific training data. We evaluate our method on out-of-domain data to explore its generalization. Despite expected variations in performance, the proposed method achieves promising results on unseen out-of-domain data when trained on the related source domain. Overall, the method can be used in out-of-domain settings, while caution should be exercised when training on significantly different domains.

  Our experiments were conducted using 7--9B parameter models, due to limitations in our available computational resources. Nevertheless, given the similar architectures and training procedures across model scales, we believe that our proposed method can be effectively applied to much larger-scale LLMs.

\section*{Ethical Considerations}
  We considered open-weights LLMs and datasets not aimed at harmful content. However, LLMs may generate potentially damaging texts for various groups of people. Uncertainty quantification techniques can help create more reliable use of neural networks. Moreover, they can be applied to detecting harmful generation, but this is not the target of this paper. Moreover, despite our proposed method demonstrating sizable performance improvements, it can still mistakenly highlight correct and innocuous generated text with high uncertainty in some cases. Thus, as with other uncertainty quantification methods, it has limited applicability. 
  
  We used writing assistants when working on this paper, in order to improve grammatical accuracy.

\section*{Acknowledgments}
  We sincerely thank the anonymous reviewers for their insightful comments and suggestions, which have greatly strengthened this paper.

\bibliography{custom}

\appendix

\clearpage
\onecolumn

\section{Additional Experimental Results}
\label{sec:appendix}

\subsection{Comparison with other UQ Methods}
\label{sec:appendix_main_results}
  Here, we present the main results for Gemma and Qwen.

  \begin{table*}[!ht] \resizebox{\textwidth}{!}{
\begin{tabular}{l|c|c|c|c|c|c|c|c|c|c|c|c|c}
\toprule
\multirow{2}{*}{\textbf{UQ Method}} & \textbf{XSum} & \textbf{SamSum} & \textbf{CNN} & \textbf{WMT19} & \textbf{MedQUAD} & \textbf{TruthfulQA} & \textbf{CoQA} & \textbf{SciQ} & \textbf{TriviaQA} & \textbf{MMLU} & \textbf{GSM8k} & \multirow{2}{*}{\multirowcell{\textbf{Mean} \\ \textbf{PRR}}} & \multirow{2}{*}{\multirowcell{\textbf{Mean} \\ \textbf{Rank}}} \\ \cline{2-12}
    & \textbf{AlignScore}  & \textbf{AlignScore} & \textbf{AlignScore} & \textbf{Comet} & \textbf{AlignScore} & \textbf{AlignScore} & \textbf{AlignScore} & \textbf{AlignScore} & \textbf{AlignScore} & \textbf{Acc.} & \textbf{Acc.} &  \\\midrule
MSP & \large\cellcolor[rgb]{0.9418435698882353,0.9280538589764706,0.9201288350882354} .321 & \large\cellcolor[rgb]{0.8694129974705882,0.9112858109117647,0.9841305319117647} .125 & \large\cellcolor[rgb]{0.9261890675039215,0.6732459732470588,0.6401732343490196} .348 & \large\cellcolor[rgb]{0.9824556940686274,0.8200795390294118,0.7599027993529412} .484 & \large\cellcolor[rgb]{0.6449978024117646,0.6934178380941176,0.9144631795921568} .004 & \large\cellcolor[rgb]{0.7581301512705882,0.8272488052941176,0.9932036999058824} .126 & \large\cellcolor[rgb]{0.9720272867117647,0.7765767393745098,0.7177742451568627} .310 & \large\cellcolor[rgb]{0.9802450306647059,0.8081382119705882,0.7477333002470589} .501 & \large\cellcolor[rgb]{0.9053078374137256,0.6343985308588236,0.6177138057666667} .649 & \large\cellcolor[rgb]{0.9077541933039216,0.6388904935882354,0.6201467828333334} .599 & \large\cellcolor[rgb]{0.9068462909411765,0.9271409192745098,0.959240051254902} .310 & \large\cellcolor[rgb]{0.9196757213862745,0.930583412827451,0.9472468817078432} .343 & \large\cellcolor[rgb]{0.9479408841470589,0.9249530282941176,0.9117495381470588} 10.27 \\
\bottomrule
\end{tabular}
}\caption{\label{tab:gemma_main_results} PRR$\uparrow$ for Gemma-2 9b model for various tasks for the considered sequence-level methods. Warmer color indicates better results. The best method is in \textbf{bold}, the second best is \underline{underlined}.}\end{table*}

  \begin{table*}[!ht] \resizebox{\textwidth}{!}{\begin{tabular}{l|c|c|c|c|c|c|c|c|c|c|c|c|c}
\toprule
\multirow{2}{*}{\textbf{UQ Method}} & \textbf{XSum} & \textbf{SamSum} & \textbf{CNN} & \textbf{WMT19} & \textbf{MedQUAD} & \textbf{TruthfulQA} & \textbf{CoQA} & \textbf{SciQ} & \textbf{TriviaQA} & \textbf{MMLU} & \textbf{GSM8k} & \multirow{2}{*}{\multirowcell{\textbf{Mean} \\ \textbf{PRR}}} & \multirow{2}{*}{\multirowcell{\textbf{Mean} \\ \textbf{Rank}}} \\ \cline{2-12}
    & \textbf{AlignScore} & \textbf{AlignScore} & \textbf{AlignScore} & \textbf{Comet} & \textbf{AlignScore} & \textbf{AlignScore} & \textbf{AlignScore} & \textbf{AlignScore} & \textbf{AlignScore} & \textbf{Acc.} & \textbf{Acc.} &  \\\midrule

MSP & \large\cellcolor[rgb]{0.7717200969960785,0.8400012947058824,0.9965252584941177} .077 & \large\cellcolor[rgb]{0.6741616707058824,0.7328555732549019,0.9441730814705882} .012 & \large\cellcolor[rgb]{0.9756268974411765,0.7893996947941176,0.729703902882353} .339 & \large\cellcolor[rgb]{0.9683898066058824,0.766374750054902,0.7090466697431372} .451 & \large\cellcolor[rgb]{0.6286169883529411,0.6698302759882353,0.8948307002647058} .030 & \large\cellcolor[rgb]{0.6286169883529411,0.6698302759882353,0.8948307002647058} -.088 & \large\cellcolor[rgb]{0.9834610525,0.8259843225764706,0.7659985578117647} .291 & \large\cellcolor[rgb]{0.9622044108411765,0.7492947789176471,0.6945295061352941} .551 & \large\cellcolor[rgb]{0.9696268857588235,0.769790744282353,0.7119501024647059} .610 & \large\cellcolor[rgb]{0.9830083599196078,0.8230648707941176,0.7629451741294118} .654 & \large\cellcolor[rgb]{0.9003004236470589,0.9251791607803921,0.9650037801960785} .268 & \large\cellcolor[rgb]{0.7446232039529412,0.8137680277960784,0.9884477548843138} .291 & \large\cellcolor[rgb]{0.7907430740941177,0.8567252977647059,0.9991571764705882} 12.91 \\
\bottomrule
\end{tabular}
}\caption{\label{tab:qwen_main_results} PRR$\uparrow$ for Qwen-2.5 7b model for various tasks for the considered sequence-level methods. Warmer color indicates better results. The best method is in \textbf{bold}, the second best is \underline{underlined}.}\end{table*}

\subsection{Results Using the ROC-AUC Metric}
\label{sec:appendix_roc_auc}
  The results with the ROC-AUC metric are presented in \Cref{tab:llama_roc_auc_results,tab:gemma_roc_auc_results,tab:qwen_roc_auc_results}. We obtain discrete versions of the generation quality metrics by thresholding the original continuous values. The thresholds were empirically determined as 0.3 for XSum, SamSum, and CNN/DailyMail; 0.5 for MedQUAD, TruthfulQA, CoQA, SciQ, and TriviaQA; and 0.85 for WMT19. The results align with the trends observed in the PRR metric. Overall, TAD outperforms the second-best method (Sheeps) by 2.4\% for LLaMa-3.1 8B, and LookBackLens by 0.1\% for Gemma-2 9B, and 2\% for Qwen-2.5 7B on average across all datasets. 

  \begin{table*}[!ht] \resizebox{\textwidth}{!}{
\begin{tabular}{l|c|c|c|c|c|c|c|c|c|c|c|c|c}
\toprule
\multirow{2}{*}{\textbf{UQ Method}} & \textbf{XSum} & \textbf{SamSum} & \textbf{CNN} & \textbf{WMT19} & \textbf{MedQUAD} & \textbf{TruthfulQA} & \textbf{CoQA} & \textbf{SciQ} & \textbf{TriviaQA} & \textbf{MMLU} & \textbf{GSM8k} & \multirow{2}{*}{\multirowcell{\textbf{Mean} \\ \textbf{ROC-AUC}}} & \multirow{2}{*}{\multirowcell{\textbf{Mean} \\ \textbf{Rank}}} \\ \cline{2-12}
    & \textbf{AlignScore} & \textbf{AlignScore} & \textbf{AlignScore} & \textbf{Comet} & \textbf{AlignScore} & \textbf{AlignScore} & \textbf{AlignScore} & \textbf{AlignScore} & \textbf{AlignScore} & \textbf{Acc.} & \textbf{Acc.} &  \\\midrule
MSP & \large\cellcolor[rgb]{0.9781854635254902,0.8875721630666666,0.8432079741549019} .705 & \large\cellcolor[rgb]{0.6996157421568627,0.7642642360784314,0.9642295513921568} .547 & \large\cellcolor[rgb]{0.9841016995,0.8604220499999999,0.8061464956666666} .669 & \large\cellcolor[rgb]{0.9704394715,0.9027982014117647,0.8675832782352941} .726 & \large\cellcolor[rgb]{0.9398111318019609,0.9290874692039215,0.9229219340686274} .717 & \large\cellcolor[rgb]{0.8980319349294118,0.9243466028941176,0.9667357341529412} .612 & \large\cellcolor[rgb]{0.9772829111803922,0.8895272809333333,0.8462654299039216} .655 & \large\cellcolor[rgb]{0.9613407263117647,0.9142840690588235,0.8885271948941176} .776 & \large\cellcolor[rgb]{0.9844312501823529,0.8554192419058824,0.7999502522156863} .809 & \large\cellcolor[rgb]{0.9717157648333333,0.9011381268078431,0.8645857989568628} .771 & \large\cellcolor[rgb]{0.9003004236470589,0.9251791607803921,0.9650037801960785} .672 & \large\cellcolor[rgb]{0.8792694128431373,0.9163932970294117,0.9792039273627451} .696 & \large\cellcolor[rgb]{0.8743412051568626,0.9138395539705882,0.9816672296372548} 11.27 \\
\bottomrule
\end{tabular}
}\caption{\label{tab:llama_roc_auc_results} ROC-AUC$\uparrow$ of UQ methods for the Llama-3.1 8b model. Warmer color indicates better results. The best method is in \textbf{bold}, the second best is \underline{underlined}.}\end{table*}

  \begin{table*}[!ht] \resizebox{\textwidth}{!}{\begin{tabular}{l|c|c|c|c|c|c|c|c|c|c|c|c|c}
\toprule
\multirow{2}{*}{\textbf{UQ Method}} & \textbf{XSum} & \textbf{SamSum} & \textbf{CNN} & \textbf{WMT19} & \textbf{MedQUAD} & \textbf{TruthfulQA} & \textbf{CoQA} & \textbf{SciQ} & \textbf{TriviaQA} & \textbf{MMLU} & \textbf{GSM8k} & \multirow{2}{*}{\multirowcell{\textbf{Mean} \\ \textbf{ROC-AUC}}} & \multirow{2}{*}{\multirowcell{\textbf{Mean} \\ \textbf{Rank}}} \\ \cline{2-12}
    & \textbf{AlignScore} & \textbf{AlignScore} & \textbf{AlignScore} & \textbf{Comet} & \textbf{AlignScore} & \textbf{AlignScore} & \textbf{AlignScore} & \textbf{AlignScore} & \textbf{AlignScore} & \textbf{Acc.} & \textbf{Acc.} &  \\\midrule

MSP & \large\cellcolor[rgb]{0.9691631781666666,0.9044582760156863,0.8705807575137254} .712 & \large\cellcolor[rgb]{0.8594926464901961,0.905996446872549,0.9888280806960784} .563 & \large\cellcolor[rgb]{0.9077541933039216,0.6388904935882354,0.6201467828333334} .681 & \large\cellcolor[rgb]{0.9818859091411765,0.8745427552862746,0.8247100216450981} .732 & \large\cellcolor[rgb]{0.9836582578333333,0.8287354144039216,0.7690800753647059} .798 & \large\cellcolor[rgb]{0.7527113214117647,0.8219973367843137,0.9915787156372549} .550 & \large\cellcolor[rgb]{0.9591408362921569,0.742086736090196,0.6888969625352941} .698 & \large\cellcolor[rgb]{0.9837721488176471,0.8654248580941176,0.812342739117647} .786 & \large\cellcolor[rgb]{0.8844643114450981,0.5946066788549021,0.5989826965745099} .863 & \large\cellcolor[rgb]{0.8898725387078432,0.6051525891921569,0.6035518578607844} .846 & \large\cellcolor[rgb]{0.9216790870960784,0.9309098270078431,0.945008558445098} .681 & \large\cellcolor[rgb]{0.9596879944529412,0.9156363617647059,0.8914368152235295} .719 & \large\cellcolor[rgb]{0.9704394715,0.9027982014117647,0.8675832782352941} 9.82 \\
\bottomrule
\end{tabular}
}\caption{\label{tab:gemma_roc_auc_results} ROC-AUC$\uparrow$ for Gemma-2 9b model for various tasks for the considered sequence-level methods. Warmer color indicates better results. The best method is in \textbf{bold}, the second best is \underline{underlined}.}\end{table*}

  \begin{table*}[!ht] \resizebox{\textwidth}{!}{\begin{tabular}{l|c|c|c|c|c|c|c|c|c|c|c|c|c}
\toprule
\multirow{2}{*}{\textbf{UQ Method}} & \textbf{XSum} & \textbf{SamSum} & \textbf{CNN} & \textbf{WMT19} & \textbf{MedQUAD} & \textbf{TruthfulQA} & \textbf{CoQA} & \textbf{SciQ} & \textbf{TriviaQA} & \textbf{MMLU} & \textbf{GSM8k} & \multirow{2}{*}{\multirowcell{\textbf{Mean} \\ \textbf{ROC-AUC}}} & \multirow{2}{*}{\multirowcell{\textbf{Mean} \\ \textbf{Rank}}} \\ \cline{2-12}
    & \textbf{AlignScore} & \textbf{AlignScore} & \textbf{AlignScore} & \textbf{Comet} & \textbf{AlignScore} & \textbf{AlignScore} & \textbf{AlignScore} & \textbf{AlignScore} & \textbf{AlignScore} & \textbf{Acc.} & \textbf{Acc.} &  \\\midrule

MSP & \large\cellcolor[rgb]{0.707400451427451,0.7734367635137256,0.9695437661686275} .524 & \large\cellcolor[rgb]{0.61490285,0.649358983,0.8768415764999999} .443 & \large\cellcolor[rgb]{0.9834610525,0.8259843225764706,0.7659985578117647} .664 & \large\cellcolor[rgb]{0.9765268001235294,0.7926054336490196,0.7326863173137255} .733 & \large\cellcolor[rgb]{0.9563825307352941,0.9183409471764705,0.8972560558823529} .670 & \large\cellcolor[rgb]{0.6355521478235294,0.6800053309882352,0.9035475637176471} .463 & \large\cellcolor[rgb]{0.9830083599196078,0.8230648707941176,0.7629451741294118} .678 & \large\cellcolor[rgb]{0.9842664748411765,0.8579206459529412,0.8030483739411765} .797 & \large\cellcolor[rgb]{0.9791396989627451,0.8021675484411765,0.7416485506941177} .819 & \large\cellcolor[rgb]{0.9807976965156863,0.8111235437352942,0.7507756750235295} .813 & \large\cellcolor[rgb]{0.9563825307352941,0.9183409471764705,0.8972560558823529} .670 & \large\cellcolor[rgb]{0.7771559349568626,0.8450485056078432,0.9977577741176471} .661 & \large\cellcolor[rgb]{0.7635661295607843,0.8323498010588235,0.9945323233411765} 13.55 \\
\bottomrule
\end{tabular}
}\caption{\label{tab:qwen_roc_auc_results} ROC-AUC$\uparrow$ for Qwen-2.5 7b model for various tasks for the considered sequence-level methods. Warmer color indicates better results. The best method is in \textbf{bold}, the second best is \underline{underlined}.}\end{table*}

\newpage
\subsection{Generalization to Out-of-Domain Tasks}
\label{sec:appendix_generalization}
  In this experiment, we examine how our approach can be generalized on the unseen datasets. For each target dataset, we construct a general QA training dataset by sampling 300 instances from the training datasets from each of other QA datasets. Thus, we evaluate TAD that is not trained on the target dataset. We conduct experiments on one dataset from each task: XSum, SamSum, CNN, WMT19, CoQA, SciQ, TriviaQA, MMLU, and GSM8k. We compare the results with three baseline methods: SAR, Semantic Density, and MSP. 

  \Cref{tab:llama_generalization_qa_results} presents the performance of the supervised methods against the MSP baseline on QA tasks, while \Cref{tab:llama_generalization_mt_ats_results} presents the results when trained on QA datasets and evaluated on summarization and translation tasks. The results demonstrate that TAD consistently outperforms baselines on unseen QA domains, while its generalization across diverse task types remains limited.
  
  \begin{table}[!ht] \centering\resizebox{0.5\textwidth}{!}{\begin{tabular}{l|c|c|c|c|c}
\toprule
\multirow{2}{*}{\textbf{UQ Method}} & \textbf{XSum} & \textbf{SamSum} & \textbf{CNN} & \textbf{WMT19} & \multirow{2}{*}{\multirowcell{\textbf{Mean} \\ \textbf{PRR}}} \\ \cline{2-5}
    & \textbf{AlignScore} & \textbf{AlignScore} & \textbf{AlignScore} & \textbf{Comet} &  \\
\midrule

MSP & \large\cellcolor[rgb]{0.9261890675039215,0.6732459732470588,0.6401732343490196} \underline{.303} & \large\cellcolor[rgb]{0.8620206859411765,0.9074551963235293,0.9878254853235294} .107 & \large\cellcolor[rgb]{0.9807976965156863,0.8111235437352942,0.7507756750235295} \underline{.329} & \large\cellcolor[rgb]{0.852836579,0.50777808,0.575116406} \textbf{.459} & \large\cellcolor[rgb]{0.9028614815235294,0.6299065681294118,0.6152808287} \underline{.299} \\
SAR & \large\cellcolor[rgb]{0.8204138911862745,0.8803757532058823,0.9989228874411764} .052 & \large\cellcolor[rgb]{0.9747269947588235,0.7861939559392157,0.7267214884509805} .166 & \large\cellcolor[rgb]{0.7825907906117646,0.8497192224705882,0.9983175350588236} .049 & \large\cellcolor[rgb]{0.8871684250764706,0.5998796340235294,0.6012672772176471} \underline{.435} & \large\cellcolor[rgb]{0.9627817115,0.9127586490352941,0.8855681539058824} .176 \\
Semantic Density & \large\cellcolor[rgb]{0.9640580048333334,0.9110985744313725,0.882570674627451} .163 & \large\cellcolor[rgb]{0.9196757213862745,0.930583412827451,0.9472468817078432} .122 & \large\cellcolor[rgb]{0.8415278407803921,0.8950213134450979,0.9948842140921569} .100 & \large\cellcolor[rgb]{0.9839369241588236,0.8629234540470588,0.8092446173921568} .295 & \large\cellcolor[rgb]{0.9563825307352941,0.9183409471764705,0.8972560558823529} .170 \\\midrule
Factoscope & \large\cellcolor[rgb]{0.9046643351764705,0.9264869997764706,0.9611612942352941} .110 & \large\cellcolor[rgb]{0.61490285,0.649358983,0.8768415764999999} .051 & \large\cellcolor[rgb]{0.6449978024117646,0.6934178380941176,0.9144631795921568} -.072 & \large\cellcolor[rgb]{0.7608481404156863,0.8297993031764705,0.9938680116235294} .083 & \large\cellcolor[rgb]{0.7232153212078432,0.7911782132705882,0.9786866527431373} .043 \\
SAPLMA & \large\cellcolor[rgb]{0.661859207627451,0.7165792202745098,0.9323611776862746} -.050 & \large\cellcolor[rgb]{0.6171885397254901,0.652770865164706,0.8798397637941177} .052 & \large\cellcolor[rgb]{0.6842533066588236,0.7455705968156863,0.9526215641058824} -.036 & \large\cellcolor[rgb]{0.61490285,0.649358983,0.8768415764999999} -.029 & \large\cellcolor[rgb]{0.61490285,0.649358983,0.8768415764999999} -.016 \\
MIND & \large\cellcolor[rgb]{0.61490285,0.649358983,0.8768415764999999} -.086 & \large\cellcolor[rgb]{0.9218513348607843,0.6650340647294117,0.635032405109804} \underline{.185} & \large\cellcolor[rgb]{0.7988883877470588,0.8636648624411765,0.9998883658882353} .064 & \large\cellcolor[rgb]{0.866948893627451,0.9100089393823529,0.9853621830490196} .158 & \large\cellcolor[rgb]{0.7988883877470588,0.8636648624411765,0.9998883658882353} .080 \\
Sheeps & \large\cellcolor[rgb]{0.9068462909411765,0.9271409192745098,0.959240051254902} .111 & \large\cellcolor[rgb]{0.8204138911862745,0.8803757532058823,0.9989228874411764} .098 & \large\cellcolor[rgb]{0.6497206297058824,0.7001240916470588,0.9199209875294118} -.067 & \large\cellcolor[rgb]{0.6667452396901961,0.7231326097019608,0.9372260848549019} .013 & \large\cellcolor[rgb]{0.7152534441254902,0.7824413707294118,0.974444709590196} .039 \\
LookBackLens & \large\cellcolor[rgb]{0.9653342981666666,0.909438499827451,0.8795731953490196} .165 & \large\cellcolor[rgb]{0.852836579,0.50777808,0.575116406} \textbf{.201} & \large\cellcolor[rgb]{0.7311771982901961,0.7999150558117647,0.9829285958960784} .005 & \large\cellcolor[rgb]{0.6286169883529411,0.6698302759882353,0.8948307002647058} -.018 & \large\cellcolor[rgb]{0.815044265017647,0.8762581198529411,0.9992540061705882} .088 \\
SATRMD & \large\cellcolor[rgb]{0.852836579,0.50777808,0.575116406} \textbf{.352} & \large\cellcolor[rgb]{0.8096589725941177,0.8720603673823529,0.9994654594098039} .096 & \large\cellcolor[rgb]{0.852836579,0.50777808,0.575116406} \textbf{.482} & \large\cellcolor[rgb]{0.9634414899941177,0.752710773145098,0.6974329388568627} .364 & \large\cellcolor[rgb]{0.852836579,0.50777808,0.575116406} \textbf{.323} \\\midrule
TAD & \large\cellcolor[rgb]{0.9696268857588235,0.769790744282353,0.7119501024647059} .259 & \large\cellcolor[rgb]{0.9261890675039215,0.6732459732470588,0.6401732343490196} .184 & \large\cellcolor[rgb]{0.61490285,0.649358983,0.8768415764999999} -.103 & \large\cellcolor[rgb]{0.7690021078509803,0.8374507968235294,0.9958609467764705} .087 & \large\cellcolor[rgb]{0.8517934440431373,0.9012928182607842,0.9914235664372548} .107 \\

\bottomrule
\end{tabular}
}\caption{\label{tab:llama_generalization_mt_ats_results} PRR$\uparrow$ for Llama-3.1 8b model for summarization and translation tasks for the considered supervised sequence-level methods trained on the general QA dataset. Unsupervised methods are not included as their performance is not dependent of the training data. Warmer colors indicate better results. The best method is in \textbf{bold}, and the second best one is \underline{underlined}.}\end{table}

\subsection{Replacing Attention Weights with Layer Integrated Gradients (LIG) Features in TAD}
\label{sec:appendix_lig}
  In this part, we expand our experiments by incorporating the use of Layer Integrated Gradients (LIG: ~\citet{lig}) as an alternative or addition to attention weights in the TAD method. The LIG features were computed using Captum's~\cite{kokhlikyan2020captum} \texttt{attribute} method, where for each predicted token $y_i$, attributions were calculated with respect to the input and previously generated tokens. Attribution vectors were aggregated across all layers and aligned to match the shape of the attention matrices.

  The motivation behind this experiment was to assess whether attribution-based interpretability features, such as LIG, which estimate token importance with respect to model outputs, could serve as a more semantically grounded alternative to raw attention weights. Given the increasing critique of attention as explanation, it was natural to test whether LIG-based representations improve uncertainty modeling.

  \Cref{tab:llama_lig_res_reordered} compares the original TAD method with two modified variants: \emph{TAD (LIG)}, which replaces attention weights entirely with LIG attributions, and \emph{TAD (MIX)}, which concatenates LIG attributions with the original attention weights. The results demonstrate that the \emph{TAD (LIG)} method performs the worst across all tasks, particularly on TruthfulQA and SamSum, where it achieves notably low PRR scores. While \emph{TAD (MIX)} significantly outperforms the LIG-only variant, the original TAD method remains superior, achieving the highest average performance across all datasets.

  The experiment demonstrates that LIG attributions, while interpretable and semantically grounded, are ineffective as a replacement for attention weights for uncertainty quantification. Furthermore, combining attention weights with LIG attributions can worsen the performance of the TAD method.

  \begin{table*}[h] 
\centering
\resizebox{0.6\textwidth}{!}{\begin{tabular}{l|c|c|c|c|c|c}
\toprule
\multirow{2}{*}{\textbf{UQ Method}} & \textbf{SamSum} & \textbf{TruthfulQA} & \textbf{CoQA} & \textbf{SciQ} & \textbf{TriviaQA} & \textbf{MMLU} \\ \cline{2-7}
    & \textbf{AlignScore} & \textbf{AlignScore} & \textbf{AlignScore} & \textbf{AlignScore} & \textbf{AlignScore} & \textbf{Acc.} \\ \midrule
TAD (LIG)    & 0.246 & .252 & 0.447 & 0.553 & 0.669 & 0.729 \\
TAD (MIX)    & \underline{0.392} & \underline{.521} & \textbf{0.510} & \underline{0.633} & \underline{0.716} & \underline{0.789} \\
TAD & \textbf{0.431} & \textbf{.565} & \underline{0.509} & \textbf{0.644} & \textbf{0.737} & \textbf{0.806} \\
\bottomrule
\end{tabular}
}\caption{\label{tab:llama_lig_res_reordered}PRR$\uparrow$ for Llama-3.1 8b model for various modifications of the TAD method using the LIG features. The best method is in \textbf{bold}, the second best is \underline{underlined}.}
\end{table*}

\subsection{Ablation Studies}
\label{sec:appendix_ablation}
  Here, we present ablation studies for various numbers of the preceding tokens, different features, and the impact of various layers for the TAD method.

  \begin{table*}[!ht] \resizebox{\textwidth}{!}{\begin{tabular}{ll|c|c|c|c|c|c|c|c|c|c|c|c}
\toprule
\multirow{2}{*}{\textbf{UQ Method}} & \multirow{2}{*}{\textbf{Aggregation}} & \textbf{XSum} & \textbf{SamSum} & \textbf{CNN} & \textbf{WMT19} & \textbf{MedQUAD} & \textbf{TruthfulQA} & \textbf{CoQA} & \textbf{SciQ} & \textbf{TriviaQA} & \textbf{MMLU} & \textbf{GSM8k} & \multirow{2}{*}{\multirowcell{\textbf{Mean} \\ \textbf{PRR}}} \\ \cline{3-13}
    & & \textbf{AlignScore} & \textbf{AlignScore} & \textbf{AlignScore} & \textbf{Comet} & \textbf{AlignScore} & \textbf{AlignScore} & \textbf{AlignScore} & \textbf{AlignScore} & \textbf{AlignScore} & \textbf{Acc.} & \textbf{Acc.} &  \\\midrule
% \midrule
TAD (LinReg) & $\frac{1}{K}\sum_{k=1}^K p_k$ & \large\cellcolor[rgb]{0.852836579,0.50777808,0.575116406} \textbf{.550} & \large\cellcolor[rgb]{0.852836579,0.50777808,0.575116406} \textbf{.535} & \large\cellcolor[rgb]{0.9837721488176471,0.8654248580941176,0.812342739117647} .444 & \large\cellcolor[rgb]{0.852836579,0.50777808,0.575116406} \textbf{.592} & \large\cellcolor[rgb]{0.852836579,0.50777808,0.575116406} \textbf{.624} & \large\cellcolor[rgb]{0.852836579,0.50777808,0.575116406} \textbf{.463} & \large\cellcolor[rgb]{0.852836579,0.50777808,0.575116406} \textbf{.392} & \large\cellcolor[rgb]{0.9765268001235294,0.7926054336490196,0.7326863173137255} \underline{.488} & \large\cellcolor[rgb]{0.8925766523392157,0.6104255443607843,0.6058364385039215} \underline{.632} & \large\cellcolor[rgb]{0.61490285,0.649358983,0.8768415764999999} .696 & \large\cellcolor[rgb]{0.852836579,0.50777808,0.575116406} \textbf{.557} & \large\cellcolor[rgb]{0.852836579,0.50777808,0.575116406} \textbf{.543} \\
TAD (LinReg) & $\sum_{k=1}^K \log p_k$ & \large\cellcolor[rgb]{0.61490285,0.649358983,0.8768415764999999} .438 & \large\cellcolor[rgb]{0.61490285,0.649358983,0.8768415764999999} .208 & \large\cellcolor[rgb]{0.61490285,0.649358983,0.8768415764999999} .422 & \large\cellcolor[rgb]{0.7022106452470588,0.7673217452235295,0.966000956317647} .358 & \large\cellcolor[rgb]{0.8718771013137254,0.9125626824411764,0.9828988807745098} .444 & \large\cellcolor[rgb]{0.6449978024117646,0.6934178380941176,0.9144631795921568} .311 & \large\cellcolor[rgb]{0.7853082184392157,0.8520545809019608,0.9985974155294117} .287 & \large\cellcolor[rgb]{0.61490285,0.649358983,0.8768415764999999} .474 & \large\cellcolor[rgb]{0.61490285,0.649358983,0.8768415764999999} .604 & \large\cellcolor[rgb]{0.9808223691882353,0.8790145912705882,0.830891189582353} \underline{.724} & \large\cellcolor[rgb]{0.61490285,0.649358983,0.8768415764999999} .355 & \large\cellcolor[rgb]{0.61490285,0.649358983,0.8768415764999999} .420 \\
TAD (MLP) & $\frac{1}{K}\sum_{k=1}^K p_k$ & \large\cellcolor[rgb]{0.9240202011823531,0.6691400189882354,0.6376028197294119} \underline{.538} & \large\cellcolor[rgb]{0.864597965917647,0.5433394684235294,0.5836202017019608} \underline{.529} & \large\cellcolor[rgb]{0.9844470791666666,0.8397397817137255,0.7814061455764706} \underline{.445} & \large\cellcolor[rgb]{0.8871684250764706,0.5998796340235294,0.6012672772176471} \underline{.578} & \large\cellcolor[rgb]{0.9842664748411765,0.8579206459529412,0.8030483739411765} \underline{.526} & \large\cellcolor[rgb]{0.864597965917647,0.5433394684235294,0.5836202017019608} \underline{.460} & \large\cellcolor[rgb]{0.8734190061058824,0.5700105097411765,0.5899980484784314} \underline{.388} & \large\cellcolor[rgb]{0.7365350864941176,0.8055387188078431,0.9853167941313725} .477 & \large\cellcolor[rgb]{0.852836579,0.50777808,0.575116406} \textbf{.634} & \large\cellcolor[rgb]{0.9296925499352942,0.9322154837294118,0.9360552653941177} .717 & \large\cellcolor[rgb]{0.9196824685392158,0.6609281104705882,0.632461990490196} \underline{.537} & \large\cellcolor[rgb]{0.9218513348607843,0.6650340647294117,0.635032405109804} \underline{.530} \\
TAD (MLP) & $\sum_{k=1}^K \log p_k$ & \large\cellcolor[rgb]{0.9236824528058823,0.9312362411882353,0.942770235182353} .492 & \large\cellcolor[rgb]{0.9133921582352942,0.9291026777686274,0.9534763223137255} .359 & \large\cellcolor[rgb]{0.852836579,0.50777808,0.575116406} \textbf{.456} & \large\cellcolor[rgb]{0.61490285,0.649358983,0.8768415764999999} .318 & \large\cellcolor[rgb]{0.61490285,0.649358983,0.8768415764999999} .328 & \large\cellcolor[rgb]{0.61490285,0.649358983,0.8768415764999999} .302 & \large\cellcolor[rgb]{0.61490285,0.649358983,0.8768415764999999} .250 & \large\cellcolor[rgb]{0.852836579,0.50777808,0.575116406} \textbf{.492} & \large\cellcolor[rgb]{0.8620206859411765,0.9074551963235293,0.9878254853235294} .615 & \large\cellcolor[rgb]{0.852836579,0.50777808,0.575116406} \textbf{.740} & \large\cellcolor[rgb]{0.8230564053823529,0.8822182482647059,0.9984342312529412} .420 & \large\cellcolor[rgb]{0.6817303976705882,0.7423918409254902,0.9505094434470589} .434 \\
\bottomrule
\end{tabular}
}\caption{\label{tab:llama_models_results} 
Comparison of various considered regression models and aggregation strategies for TAD (PRR$\uparrow$, Llama-3.1 8b model). Warmer colors indicate better results.The best method is in \textbf{bold}, the second best is \underline{underlined}.}\end{table*}
  \begin{table*}[h] 
\centering
\resizebox{0.75\textwidth}{!}{\begin{tabular}{l|c|c|c|c|c|c|c|c}
\toprule
\multirow{2}{*}{\textbf{UQ Method}} & \textbf{XSum} & \textbf{SamSum} & \textbf{CNN} & \textbf{WMT19} & \textbf{MedQUAD} & \textbf{TruthfulQA} & \textbf{GSM8k} & \multirow{2}{*}{\multirowcell{\textbf{Mean} \\ \textbf{PRR}}} \\ \cline{2-8}
    & \textbf{AlignScore} & \textbf{AlignScore} & \textbf{AlignScore} & \textbf{Comet} & \textbf{AlignScore} & \textbf{AlignScore} & \textbf{Acc.} &  \\\midrule
TAD (1 token)  & \large\cellcolor[rgb]{0.8718771013137254,0.9125626824411764,0.9828988807745098} \underline{.554} & \large\cellcolor[rgb]{0.7662841187058823,0.8349002989411765,0.9951966350588235} .538 & \large\cellcolor[rgb]{0.61490285,0.649358983,0.8768415764999999} .437 & \large\cellcolor[rgb]{0.61490285,0.649358983,0.8768415764999999} .563 & \large\cellcolor[rgb]{0.61490285,0.649358983,0.8768415764999999} .496 & \large\cellcolor[rgb]{0.61490285,0.649358983,0.8768415764999999} .407 & \large\cellcolor[rgb]{0.61490285,0.649358983,0.8768415764999999} .513 & \large\cellcolor[rgb]{0.61490285,0.649358983,0.8768415764999999} .519 \\
TAD (2 tokens)  & \large\cellcolor[rgb]{0.852836579,0.50777808,0.575116406} \textbf{.560} & \large\cellcolor[rgb]{0.9367011412764705,0.6934798195294117,0.6531662319882353} \underline{.547} & \large\cellcolor[rgb]{0.9683898066058824,0.766374750054902,0.7090466697431372} .442 & \large\cellcolor[rgb]{0.9479408841470589,0.9249530282941176,0.9117495381470588} .578 & \large\cellcolor[rgb]{0.806966326382353,0.8699614911470588,0.9995711860294118} .534 & \large\cellcolor[rgb]{0.852836579,0.50777808,0.575116406} \textbf{.467} & \large\cellcolor[rgb]{0.7662841187058823,0.8349002989411765,0.9951966350588235} .524 & \large\cellcolor[rgb]{0.9357462556294118,0.9311546896588235,0.9285081320294117} .532 \\
TAD (5 tokens) & \large\cellcolor[rgb]{0.6379135614705882,0.6833584577647058,0.9062764676862745} .550 & \large\cellcolor[rgb]{0.852836579,0.50777808,0.575116406} \textbf{.549} & \large\cellcolor[rgb]{0.9575785619705882,0.7384632635882353,0.686089707382353} \underline{.443} & \large\cellcolor[rgb]{0.9765268001235294,0.7926054336490196,0.7326863173137255} \underline{.585} & \large\cellcolor[rgb]{0.9849255762058824,0.8479150297647058,0.7906558870392157} \underline{.584} & \large\cellcolor[rgb]{0.9834812190705882,0.867835001309804,0.8154382697392157} .446 & \large\cellcolor[rgb]{0.8616576191882352,0.5344491213176471,0.5814942527764706} \underline{.556} & \large\cellcolor[rgb]{0.9765268001235294,0.7926054336490196,0.7326863173137255} \underline{.539} \\\midrule
TAD (10 tokens)  & \large\cellcolor[rgb]{0.61490285,0.649358983,0.8768415764999999} .550 & \large\cellcolor[rgb]{0.61490285,0.649358983,0.8768415764999999} .535 & \large\cellcolor[rgb]{0.852836579,0.50777808,0.575116406} \textbf{.444} & \large\cellcolor[rgb]{0.852836579,0.50777808,0.575116406} \textbf{.592} & \large\cellcolor[rgb]{0.852836579,0.50777808,0.575116406} \textbf{.624} & \large\cellcolor[rgb]{0.9028614815235294,0.6299065681294118,0.6152808287} \underline{.463} & \large\cellcolor[rgb]{0.852836579,0.50777808,0.575116406} \textbf{.557} & \large\cellcolor[rgb]{0.852836579,0.50777808,0.575116406} \textbf{.545} \\
\bottomrule
\end{tabular}
}\caption{\label{tab:llama_tokens_results} PRR$\uparrow$ for Llama-3.1 8b model for various tasks for the various number of preceding tokens for the TAD method. Warmer color indicates better results. The best method is in \textbf{bold}, the second best is \underline{underlined}.}\end{table*}

  \begin{table*}[!ht] \centering\resizebox{0.65\textwidth}{!}{\begin{tabular}{l|c|c|c|c|c|c|c|c}
\toprule
\multirow{2}{*}{\textbf{UQ Method}} & \textbf{XSum} & \textbf{SamSum} & \textbf{CNN} & \textbf{WMT19} & \textbf{MedQUAD} & \textbf{TruthfulQA} & \textbf{GSM8k} & \multirow{2}{*}{\multirowcell{\textbf{Mean} \\ \textbf{PRR}}} \\ \cline{2-8}
    & \textbf{AlignScore} & \textbf{AlignScore} & \textbf{AlignScore} & \textbf{Comet} & \textbf{AlignScore} & \textbf{AlignScore} & \textbf{Acc.} &  \\
\midrule
TAD (Sequence-level) & \large{.541} & \large\textbf{.550} & \large{.411} & \large\textbf{.640} & \large{.500} & \large{.420} & \large{.517} & \large{.511} \\
TAD & \large\textbf{.550} & \large{.535} & \large\textbf{.444} & \large{.592} & \large\textbf{.624} & \large\textbf{.463} & \large\textbf{.557} & \large\textbf{.538} \\
\bottomrule
\end{tabular}
}\caption{\label{tab:llama_taq_seq} PRR$\uparrow$ for the modifications of the TAD method for the Llama-3.1 8b model. The best method is in \textbf{bold}, the second best is {underlined}.}\end{table*}

  \begin{table*}[!ht] \resizebox{\textwidth}{!}{\begin{tabular}{l|c|c|c|c|c|c|c|c|c|c|c|c}
\toprule
\multirow{2}{*}{\textbf{UQ Method}} & \textbf{XSum} & \textbf{SamSum} & \textbf{CNN} & \textbf{WMT19} & \textbf{MedQUAD} & \textbf{TruthfulQA} & \textbf{CoQA} & \textbf{SciQ} & \textbf{TriviaQA} & \textbf{MMLU} & \textbf{GSM8k} & \multirow{2}{*}{\multirowcell{\textbf{Mean} \\ \textbf{PRR}}} \\ \cline{2-12}
    & \textbf{AlignScore} & \textbf{AlignScore} & \textbf{AlignScore} & \textbf{Comet} & \textbf{AlignScore} & \textbf{AlignScore} & \textbf{AlignScore} & \textbf{AlignScore} & \textbf{AlignScore} & \textbf{Acc.} & \textbf{Acc.} &  \\\midrule

TAD (probs.) & \large\textbf{.554} & \large\textbf{.538} & \large.437 & \large.563 & \large.496 & \large.407 & \large.385 & \large.461 & \large\underline{.628} & \large\underline{.727} & \large.513 & \large.519 \\
TAD (attention) & \large.549 & \large.532 & \large\textbf{.453} & \large\underline{.590} & \large\underline{.624} & \large\textbf{.465} & \large\textbf{.396} & \large\underline{.473} & \large.609 & \large\textbf{.730} & \large\underline{.538} & \large\underline{.542}  \\
\midrule
TAD (attention+probs.) & \large\underline{.550} & \large\underline{.535} & \large\underline{.444} & \large\textbf{.592} & \large\textbf{.624} & \large\underline{.463} & \large\underline{.392} & \large\textbf{.488} & \large\textbf{.632} & \large.724 & \large\textbf{.557} & \large\textbf{.545} \\

\bottomrule
\end{tabular}
}\caption{\label{tab:llama_ablation_results} PRR$\uparrow$ for Llama-3.1 8b model for various tasks for different features for the TAD method. Warmer color indicates better results. The best method is in \textbf{bold}, the second best is \underline{underlined}.}\end{table*}
  \begin{table*}[!ht] \resizebox{\textwidth}{!}{\begin{tabular}{l|c|c|c|c|c|c|c|c|c|c|c|c}
\toprule
\multirow{2}{*}{\textbf{UQ Method}} & \textbf{XSum} & \textbf{SamSum} & \textbf{CNN} & \textbf{WMT19} & \textbf{MedQUAD} & \textbf{TruthfulQA} & \textbf{CoQA} & \textbf{SciQ} & \textbf{TriviaQA} & \textbf{MMLU} & \textbf{GSM8k} & \multirow{2}{*}{\multirowcell{\textbf{Mean} \\ \textbf{PRR}}} \\ \cline{2-12}
    & \textbf{AlignScore} & \textbf{AlignScore} & \textbf{AlignScore} & \textbf{Comet} & \textbf{AlignScore} & \textbf{AlignScore} & \textbf{AlignScore} & \textbf{AlignScore} & \textbf{AlignScore} & \textbf{Acc.} & \textbf{Acc.} &  \\\midrule
% \midrule
TAD (1 step) & \large{.013} & \large{.153} & \large{.195} & \large{.269} & \large{-.121} & \large{.156} & \large{.257} & \large{.426} & \large{.522} & \large{.541} & \large{.205} & \large{.238} \\
TAD (2 step) & \large\textbf{.550} & \large\textbf{.535} & \large\textbf{.444} & \large\textbf{.592} & \large\textbf{.624} & \large\textbf{.463} & \large\textbf{.392} & \large\textbf{.488} & \large\textbf{.632} & \large\textbf{.724} & \large\textbf{.557} & \large\textbf{.545} \\
\bottomrule
\end{tabular}
}\caption{\label{tab:llama_steps_results} PRR$\uparrow$ for Llama-3.1 8b model for various tasks for the different number of learning steps for the TAD method. Warmer color indicates better results. The best method is in \textbf{bold}.
}
\end{table*}

  \begin{figure}[H]
    \centering
    \resizebox{0.8\textwidth}{!}{
    \includegraphics[trim={0.cm 0.cm 0.cm 0.cm},clip,width=\linewidth]{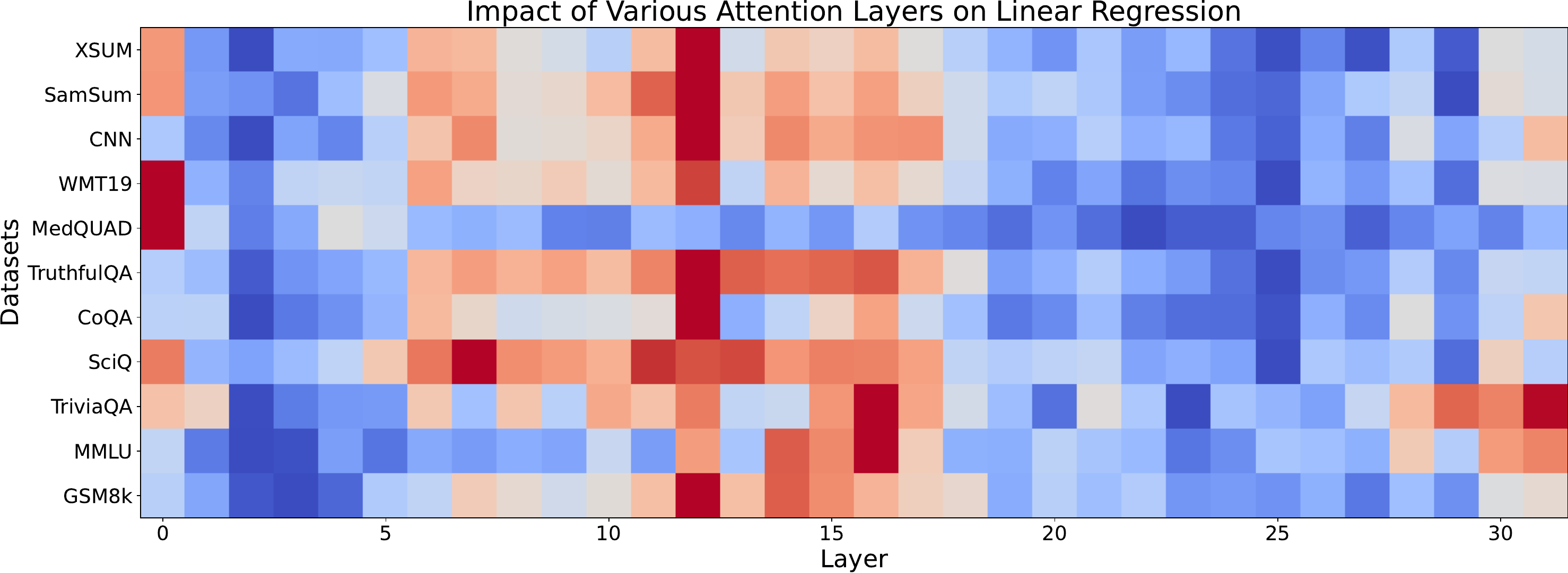}
    }
    \caption{Normalized average weights of linear regression for different attention layers in the TAD method across the considered datasets. Warmer color indicates a higher impact on the TAD performance.}
    \label{fig:attention_layers}
  \end{figure}

  \FloatBarrier

\newpage

\section{Computational Resources and Efficiency}
  All experiments were conducted on a single NVIDIA H100 GPU. On average, training a single model across all datasets took over 750 GPU hours, while inference on the test set took 260 GPU hours.

\clearpage
\section{Hyperparameters}
\subsection{Optimal Hyperparameters for TAD}
\label{sec:hp_appendix}
  The optimal hyperparameters for TAD for various considered regression models and different aggregation strategies are presented in \Cref{tab:llama_tad_hp,tab:gemma_tad_hp,tab:qwen_tad_hp} for Llama-3.1 8b, Gemma-2 9b, and Qwen-2.5 7b models respectively. These hyperparameters are obtained using cross-validation with five folds using the training dataset. We train a regression model on $k-1$ folds of the training dataset and estimate uncertainty on the remaining fold. The optimal hyperparameters are selected according to the best average PRR for AlignScore. Finally, we use these hyperparameters to train the regression model on the entire training set. 
  \vspace{0.2cm}\\
  The hyperparameter grid for the linear regression is the following: \\ 
    \textbf{L2 regularization}: [1e+1, 1, 1e-1, 1e-2, 1e-3, 1e-4].
  \vspace{0.2cm}\\
  The hyperparameter grid for the MLP is the following: \\ 
    \textbf{Num. of layers}: [2, 4]; \\
    \textbf{Num. of epochs}: [10, 20, 30]; \\ 
    \textbf{Learning rate}: [1e-5, 3e-5, 5e-5]; \\ 
    \textbf{Batch size}: [64, 128].
  For both models, we include aggregation strategies in the hyperparameter grid for the final configuration. 

  \begin{table*}[!ht] 
\centering

\resizebox{\textwidth}{!}{\begin{tabular}{ll|c|c|c|c|c|c|c|c|c|c|c}
\toprule
\textbf{UQ Method} & \textbf{Aggregation} & \textbf{XSum} & \textbf{SamSum} & \textbf{CNN} & \textbf{WMT19} & \textbf{MedQUAD} & \textbf{TruthfulQA} & \textbf{CoQA} & \textbf{SciQ} & \textbf{TriviaQA} & \textbf{MMLU} & \textbf{GSM8k} \\\midrule
TAD (MLP) & $\frac{1}{K}\sum_{k=1}^K p_k$ & 4, 30, 1e-05, 0, 128 & 4, 30, 3e-05, 0, 128 & 4, 30, 3e-05, 0, 128 & 2, 30, 3e-05, 0, 128 & 2, 30, 1e-05, 0, 128 & 4, 30, 3e-05, 0, 64 & 4, 30, 5e-05, 0, 128 & 4, 30, 3e-05, 0, 128 & 4, 30, 3e-05, 0, 128 & 4, 30, 5e-05, 0, 128 & 4, 30, 1e-05, 0, 128 \\
TAD (MLP) & $\sum_{k=1}^K \log p_k$ & 4, 30, 3e-05, 0, 128 & 4, 30, 5e-05, 0, 64 & 4, 30, 5e-05, 0, 64 & 2, 30, 3e-05, 0, 128 & 4, 30, 1e-05, 0, 128 & 4, 30, 5e-05, 0, 128 & 2, 30, 5e-05, 0, 64 & 4, 30, 5e-05, 0, 64 & 4, 30, 3e-05, 0, 128 & 4, 30, 5e-05, 0, 128 & 4, 30, 1e-05, 0, 128 \\
TAD (LinReg) & $\frac{1}{K}\sum_{k=1}^K p_k$ & 1 & 1 & 10.0 & 1 & 0.001 & 0.1 & 1 & 0.01 & 10.0 & 1 & 10.0 \\
TAD (LinReg) & $\sum_{k=1}^K \log p_k$ & 10.0 & 0.01 & 1 & 0.001 & 0.001 & 1 & 1 & 1 & 10.0 & 1 & 0.1 \\
\bottomrule
\end{tabular}
}\caption{\label{tab:llama_tad_hp} Optimal values of the hyper-parameters for the TAD methods for the Llama-3.1 8b model.}
\vspace{0.5cm}

\resizebox{\textwidth}{!}{\begin{tabular}{l|c|c|c|c|c|c|c|c|c|c|c}
\toprule
\textbf{UQ Method} & \textbf{XSum} & \textbf{SamSum} & \textbf{CNN} & \textbf{WMT19}& \textbf{MedQUAD} & \textbf{TruthfulQA} & \textbf{CoQA} & \textbf{SciQ} & \textbf{TriviaQA} & \textbf{MMLU} & \textbf{GSM8k} \\\midrule
TAD (LinReg) & 0.01 & 1 & 1 & 1 & 0.001 & 0.1 & 10.0 & 0.1 & 10.0 & 1 & 0.1 \\
\bottomrule
\end{tabular}
}\caption{\label{tab:gemma_tad_hp} Optimal values of the hyper-parameters for the final configuration of the TAD method for the Gemma-2 9b model. For CNN, SciQ, and MMLU, $\sum_{k=1}^K \log p_k$ is the best aggregation method, whereas $\frac{1}{K}\sum_{k=1}^K p_k$ performs best on all other datasets.}
\vspace{0.5cm}

\resizebox{\textwidth}{!}{\begin{tabular}{l|c|c|c|c|c|c|c|c|c|c|c}
\toprule
\textbf{UQ Method} & \textbf{XSum} & \textbf{SamSum} & \textbf{CNN} & \textbf{WMT19} & \textbf{MedQUAD} & \textbf{TruthfulQA} & \textbf{CoQA} & \textbf{SciQ} & \textbf{TriviaQA} & \textbf{MMLU} & \textbf{GSM8k} \\\midrule
TAD (LinReg) & 0.01 & 1 & 10.0 & 1 & 0.001 & 0.1 & 10.0 & 0.1 & 10.0 & 1 & 0.1 \\
\bottomrule
\end{tabular}
}\caption{\label{tab:qwen_tad_hp} Optimal values of the hyper-parameters for the final configuration of the TAD method for the Qwen-2.5 7b model. For MMLU, $\sum_{k=1}^K \log p_k$ is the best aggregation method, whereas $\frac{1}{K}\sum_{k=1}^K p_k$ performs best on all other datasets.}

\end{table*}

\subsection{LLM Generation Hyperparameters}
\label{sec:llm_hp}

  \begin{table*}[h] 
\centering\resizebox{\textwidth}{!}{
\begin{tabular}{l|c|c|c|c|c|c|c|c}
\toprule
\textbf{Dataset} & \textbf{Task} & \textbf{Max Input Length} & \textbf{Generation Length} & \textbf{Temperature} & \textbf{Top-p} & \textbf{Do Sample} & \textbf{Beams} & \textbf{Repetition Penalty} \\
\midrule
 XSum & \multirow{3}{*}{TS} & \multirow{11}{*}{-} & 56 & \multirow{11}{*}{1.0} & \multirow{11}{*}{1.0} & \multirow{11}{*}{False} & \multirow{11}{*}{1} & \multirow{11}{*}{1}  \\
 SamSum &  &  & 128 &  &  &  &  & \\%& \multirow{2}{*}{TS} & \multirow{10}{*}{-} & 128 & \multirow{10}{*}{1.0} & \multirow{10}{*}{1.0} & \multirow{10}{*}{False} & \multirow{10}{*}{1} & \multirow{10}{*}{1}  \\ 
 CNN &  &  & 128 &  &  &  &  & 1.2 \\

 WMT19 &  \multirow{1}{*}{\multirowcell{MT}} &  & 107 &  &  &  &  & \\

 % PubMedQA & \multirow{3}{*}{\multirowcell{QA \\ Long answer}} & & 128 &  &  &  &  & \\
 MedQUAD & \multirow{3}{*}{\multirowcell{QA \\ Long answer}} & & 128 &  &  &  &  & \\
 TruthfulQA & &  & 128 &  &  &  &  & \\
 GSM8k &  &  & 256 &  &  &  &  & \\

 CoQA & \multirow{3}{*}{\multirowcell{QA \\ Short answer}} & & 20 & & & & & \\
 SciQ &  &  & 20 &  &  &  &  &  \\
 TriviQA &  &  & 20 &  &  &  &  &  \\

 MMLU &  \multirow{1}{*}{\multirowcell{MCQA}} &  & 3 &  &  &  &  &  \\

\bottomrule
\end{tabular}}
\caption{\label{tab:llm_hyperparameters} Values of the text generation hyper-parameters for all LLMs used in our experiments.}
\end{table*}

\clearpage
\section{Dataset Statistics}
\label{sec:ds_stats}
  Statistics about the datasets are provided in \Cref{tab:dataset_stat}. For TS, we experiment with CNN/DailyMail~\cite{see-etal-2017-get}, XSum~\cite{narayan-etal-2018-dont}, 
  and SamSum~\cite{gliwa-etal-2019-samsum}. For the long answer QA task, we use MedQUAD~\cite{medquad}, which consists of real medical questions, TruthfulQA~\cite{lin-etal-2022-truthfulqa}, which consists of questions that some people would answer incorrectly due to a false belief or a misconception, and GSM8k~\cite{gsm8k} with a grade school math questions. 
  For the QA task with short answers, we follow previous work on UQ~\cite{kuhn2023semantic,duan2023shifting,lin2023generating} and we use three datasets: SciQ~\cite{welbl-etal-2017-crowdsourcing}, CoQA~\cite{reddy-etal-2019-coqa}, and TriviaQA~\cite{joshi-etal-2017-triviaqa}. For multiple-choice QA, we use MMLU~\cite{mmlu}, a widely used benchmark for evaluating LLMs.
  For MT, we use WMT19~\cite{barrault-etal-2019-findings}, focusing on translations from German to English.

  \begin{table}[h] \footnotesize  \centering\resizebox{0.6\textwidth}{!}{\begin{tabular}{c|l|c|c|c}
\toprule
\textbf{Task} & \textbf{Dataset} & \textbf{N-shot} & \multirowcell{\textbf{Train texts} \\ \textbf{for TAD}} & \multirowcell{\textbf{Evaluation} \\ \textbf{texts}} \\
\midrule
\multirow{3}{*}{\multirowcell{Text \\ Summarization}} & CNN/DailyMail & 0 & 500 & 1,000 \\
& XSum & 0 & 1,000 & 2,000 \\
& SamSum & 0 & 2,000 & 819 \\
\midrule
\multirow{1}{*}{\multirowcell{MT}} & WMT19 De-En & 0 & 2,000 & 2,000 \\
\midrule
% \multirow{3}{*}{\multirowcell{QA \\ Long answer}} & PubMedQA & 0 & 2,000 & 2,000 \\
\multirow{3}{*}{\multirowcell{QA \\ Long answer}} & MedQUAD & 5 & 500 & 1,000 \\
& TruthfulQA & 5 & 408 & 409 \\
& GSM8k & 5 & 700 & 1,319 \\
\midrule
\multirow{4}{*}{\multirowcell{QA \\ Short answer}} & SciQ & 0 & 2,000 & 1,000 \\
& CoQA & \multirowcell{all preceding \\ questions} & 2,000 & 2,000 \\
& TriviaQA & 5 & 2,000 & 2,000 \\
\midrule
\multirow{1}{*}{\multirowcell{MCQA}} & MMLU & 5 & 2,000 & 2,000 \\
\bottomrule
\end{tabular}
}\caption{\label{tab:dataset_stat} Statistics about the datasets used for evaluation. }
\end{table}

\newpage

\section{Generating Training Data for TAD}
\label{sec:tad_training_data}

  \begin{algorithm}[ht]
    \caption{Generating training data for TAD}\label{alg:tad_training_data}
    \KwData{Input prompt $\xv_k$, LLM generation $\yv_k = \yv_{1:n_k}$, token probabilities $p(y_i\mid \yv_{<i},\xv_k)$, number of preceding tokens $N$, vector of LLM attention weights $a_{i, i-l}$ from the $(i-l)$-th token to the $i$-th token from all layers and heads, and step of the training procedure $j$}
    \KwResult{Feature vectors $z_i^k$, $k=1\dots K$, $i=2\dots n_k$}

   \BlankLine
    \tcp{Estimate unconditional probability for the first token}
    $\hat{p}_k(y_1) = \mathrm{sim}(\yv_k, \yv_k^*)$\;

  \BlankLine
    \For{$i \gets 2$ \KwTo $n_k$}{

      \tcp{Construct token‑level features}
      \eIf{$j==1$}{
          \tcp{On the first training step, we use only probabilities as features}
          $z_i^k\gets\bigoplus_{l=1}^{\min\{N, i-1\}} \Big[ \, p(y_{i-l} \mid \yv_{<i-l}, \xv_k) \Big] \oplus \Big[ p(y_i \mid \yv_{<i}, \xv_k) \Big] $\ ;
      }{
          \tcp{On the next training steps, we use all features}
          $z_i^k\gets\bigoplus_{l=1}^{\min\{N, i-1\}} \Big[ \, p(y_{i-l} \mid \yv_{<i-l}, \xv_k), \  \hat{p}_k(y_{i-l}), \ a_{i, i-l}\, \Big] \oplus \Big[ p(y_i \mid \yv_{<i}, \xv_k) \Big]$\;
      }
      \BlankLine
      \tcp{If $N > i - 1$, we pad $z_i^k$ with zeros to ensure they have the same length}  
      \If{$i-1 < N$}{
        $z_i^k \gets z_i^k \oplus \mathbf{0}_{(2 + |a_{i, i-l}|)(N - i + 1)}$\;
      }

    \BlankLine
    \tcp{Estimate token‑level unconditional probability}
      \eIf{$j==1$}{
          \tcp{On the first training step, we use ground truth}
          $\hat{p}_k(y_i) = \mathrm{sim}(\yv_k, \yv_k^*)$\;
      }{
          \tcp{On the next training steps, we use trained function $\conf(\cdot)$}
          $\hat{p}_k(y_i) = \conf(z_i^k)$\;
      }   
    }
    \Return{$z_i^k$, $k=1\dots K$, $i=2\dots n_k$}\;
  \end{algorithm}

\end{document}